\documentclass{article}
\usepackage[letterpaper,top=2cm,bottom=2cm,left=2cm,right=1.8cm,marginparwidth=1.75cm]{geometry}

\usepackage{amssymb,amsmath,amsthm,bm}

\newcommand{\R}{\mathbb{R}}
\usepackage{authblk}
\usepackage{xcolor}
\usepackage{pifont}

\usepackage{graphicx}
\usepackage{subcaption}
\usepackage{float}
\usepackage{multirow}
\usepackage{rotating}
\usepackage{booktabs}
\usepackage[normalem]{ulem}
\usepackage{pdflscape}

\usepackage{algorithm}

\usepackage{amsmath}

\usepackage{amsmath,amssymb,bm,graphicx,booktabs}
\usepackage{algorithm}
\usepackage{algpseudocode}

\usepackage{microtype}
\usepackage{comment}
\usepackage{soul}
\usepackage{todonotes}

\newcommand{\cE}{\mathcal{E}}
\newcommand{\cD}{\mathcal{D}}
\newcommand{\cH}{\mathcal{H}}
\newcommand{\cJ}{\mathcal{J}}

\usepackage{tocloft}
\usepackage{hyperref}
\hypersetup{
    colorlinks=true,
    linkcolor=blue,
    citecolor=blue,
    urlcolor=blue
}

\title{
\Large \bfseries
\textcolor{black!35!blue}{GenVoid: Uncertainty-Aware Learning of Subsurface Material Defects with an Experimentally Validated Physics-Informed Generative Model}}
\author[]{Trishit Mondal}
\author[]{Prajwal Bharadwaj}
\author[]{Nikhil Karanjgaokar}
\author[]{Ameya D. Jagtap\thanks{Corresponding author: Ameya D. Jagtap (ajagtap@wpi.edu, ameyadjagtap@gmail.com)}}

\affil[]{\textit{\small{Aerospace Engineering Department, Worcester Polytechnic Institute, Worcester, MA 01609, USA.}}}
\date{}
\begin{document}

\maketitle

\begin{abstract}
Internal voids are ubiquitous defects in manufactured structures, yet their characterization remains challenging because their geometry is hidden and can only be inferred indirectly from accessible measurements. Here we introduce \textit{GenVoid}, a physics-informed generative model-based framework for identifying internal voids in complex two- and three-dimensional solids from surface displacement measurements alone. By incorporating the governing mechanics into a generative inference framework, \textit{GenVoid} enables void identification across linear elastic, hyperelastic and plastic material behaviors and accommodates complex two- and three-dimensional structural geometries. Importantly, the framework explicitly accounts for uncertainty and noise in displacement measurements, producing probabilistic reconstructions of internal void geometry rather than a single deterministic estimate. We demonstrate the approach using high-fidelity synthetic datasets and experimentally measured displacement fields obtained from in-situ mechanical experiments, establishing its ability to infer hidden voids from realistic displacement measurements. To quantify the fundamental limits of such inference, we further introduce an observability measure that characterizes the sensitivity of boundary measurements to localized stiffness perturbations within the interior under an ensemble of applied loads. This framework provides a direct connection between defect location, sensor configuration and reconstruction fidelity, enabling systematic assessment of how the number and spatial distribution of boundary measurements govern void-identification accuracy. To this end, these results establish a physics-informed and uncertainty-aware approach for non-invasive characterization of hidden defects and provide a quantitative basis for designing measurement strategies for inverse problems in solid mechanics.
\end{abstract}

\maketitle
\vspace{0.2cm}

 \begin{small}Keywords: \textit{Physics-informed Generative Models}; \textit{Inverse Problems}; \textit{Nondestructive Evaluation}; \textit{Void Identification}; \textit{Uncertainty Quantification}.
\end{small}

\begin{figure}[htpb]
\centering
\includegraphics[width=0.95\textwidth, trim=1cm 1cm 1cm 0cm]{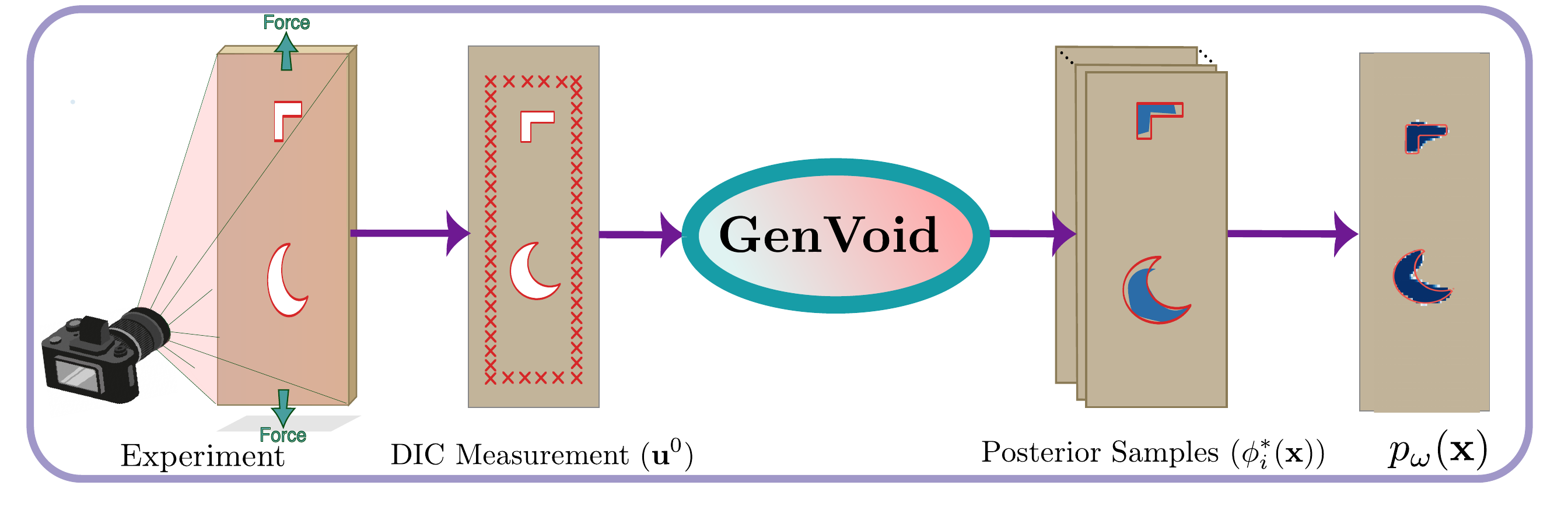}
\caption{Schematic of the experimental validation of void reconstruction using \textit{GenVoid}.} 
\label{fig:exp_pipeline}
\end{figure}
\section{Introduction}
Identification of internal voids is an important geometric identification inverse problem in continuum solid mechanics, with direct relevance to nondestructive evaluation, structural health monitoring, and the qualification and failure assessment of manufactured components. Interior voids arising from porosity during casting or additive manufacturing, as well as corrosion pits beneath coatings, can reduce stiffness, perturb load transfer, and initiate failure long before they become detectable from the exterior; see \cite{sun2023review}, for more details. Because the interior of a structure is generally inaccessible to direct instrumentation, these defects must be inferred from their influence on measurable responses at the external boundary. The inverse problem is therefore to determine the number, locations, sizes, and shapes of unknown voids from boundary measurements under prescribed mechanical loading. More generally, this problem belongs to a broad class of geometry identification problems in continuum solid mechanics, where unknown internal defects or boundaries must be characterized from indirect measurements. Unlike conventional parameter identification, the unknown in void identification is a geometry whose location, morphology, number of connected components, and topology are not known a priori, making the inverse problem intrinsically more challenging. Furthermore, the mapping from an interior void geometry to boundary measurements is strongly smoothing, so that a void located deep within a structure may produce only a weak surface response, while geometrically different void configurations may generate responses that differ by less than the measurement noise. Consequently, inverse void identification is inherently ill-posed and sensitive to measurement uncertainty, modeling errors, and the initial estimate of the unknown geometry.

Traditional computational approaches to inverse void identification predominantly rely on level-set representations; originating from the seminal work of Osher and Sethian~\cite{osher1988}, which track evolving interfaces implicitly via Hamilton--Jacobi equations to allow topological changes without explicit mesh fitting. In classical shape and topology optimization, Allaire et al.~\cite{allaire2004} combined level-set interfaces with adjoint-based shape derivatives on fixed Eulerian meshes using an ersatz-material approach, while Sukumar et al.~\cite{sukumar2001} coupled level sets with the extended finite element method (XFEM) to enrich local basis functions at defect boundaries. To manage high-dimensional design spaces, parametric level-set methods represent the interface field using a reduced set of radial basis functions (RBFs) or neural network parameters; notably, Yu et al.~\cite{yu2023} utilized compactly supported RBFs with ersatz material models and Levenberg--Marquardt optimization for inverse void recovery from boundary displacements, while Deng and To~\cite{deng2021dnnls} parametrized the level-set field directly via deep neural network weights. Despite their flexibility, these gradient-based level-set schemes remain acutely sensitive to the initial geometric configuration, as sensitivity fields often decay away from existing interfaces, rendering them incapable of discovering uninitialized voids. Parallel developments in deep learning address defect characterization across several distinct paradigms. Supervised image- and signal-classification pipelines apply convolutional and graph neural networks directly to experimental measurements for localized defect detection, such as Bayesian graph networks on strain sensor arrays~\cite{mylonas2022gnncrack}, 1D CNNs on eddy-current signals~\cite{meng2021eddycurrent}, 2D CNNs on ultrasonic, visual, and X-ray CT imagery~\cite{choi2021uavcrack,ma2025cfrpcnn,gobert2020porosity}, and deep networks for ultrasonic guided-wave diagnosis~\cite{zhang2022shmweldment}; though these require massive labeled datasets and lack a physics-based forward solver. Alternatively, physics-informed neural networks (PINNs), introduced by Raissi et al.~\cite{raissi2019pinn} and extended in many work \cite{jagtap2020extended,cpinn_intro,menon2026scientific,menon2026fekan} for different applications \cite{jagtap2022deep,menon2025anant,menon2026intelligent,mhaskar2026approximation,jagtap2022deep,jagtap2022deepD,abbasi2025history,abbasi2025challenges,jagtap2020adaptive,shukla2021physics,jagtap2022physics,jagtap2020locally,maity2026enpinn,menon2026tackling}, embed governing partial differential equations (PDEs) directly into network loss functions via automatic differentiation. PINNs have been applied to parameterize unknown void boundaries across elastic, hyperelastic, and plastic regimes~\cite{zhang2022pinn}, map continuous heterogeneous modulus fields~\cite{chen2021elastnet,zhang2020pinnelasticityimaging}, and locate voids via phase-field variables~\cite{he2025phasefieldpinn} or inverse obstacle scattering~\cite{yin2025scattering}, while networks like ElastNet~\cite{chen2021elastnet} and Fourier neural operators (FNO)~\cite{li2020fno} map full-field strains directly to stiffness properties~\cite{kim2026fnoelastography,patel2019circumventing}. Closer to biomedical elasticity imaging, Kamali et al.
\cite{kamali2023elasticity} used a PINN to recover elastic modulus and
Poisson's ratio fields simultaneously from strain data. In parallel, operator learning frameworks \cite{lu2021learning,peyvan2024riemannonets,zhang2026bubbleokan,goswami2024learning}, including Learned Primal-Dual architectures~\cite{adler2018primaldual} and neural representations for topology optimization~\cite{zehnder2021ntopo,mallon2025neurallevelset}, demonstrate that deep models can approximate complex boundary-to-geometry maps, supported by theoretical proofs for DeepONet approximations of Dirichlet-to-Neumann operators~\cite{castro2024calderon}. Furthermore, neural operators such as the FNO~\cite{li2020fno} and the Geometry-Informed Neural Operator (GINO)~\cite{li2023gino} leverage signed-distance fields and latent Fourier layers to learn mesh-invariant solution operators across irregular 3D domains. 

A parallel line of work embeds governing equations into a generative
model's training objective. PI-GANs \cite{yang2020physics} combine
generative adversarial training with automatic differentiation to
enforce the governing equations, solving forward, inverse, and mixed
problems from scattered measurements; PI-VEGAN \cite{gao2023pi} added a
variational embedding to better capture the latent distribution. Most
relevant here, Warner et al.~\cite{warner2020inverse} used a PI-GAN to
infer a 2D solid's full elastic-modulus field directly from measured
deformation, with no labeled stiffness data, by enforcing the elasticity
PDE as a residual penalty. Physics-informed VAEs pair a generative
encoder--decoder with a physics loss: PI-VAE \cite{zhong2023pi} trains
the encoder-decoder against the governing equations directly, while
p$^3$VAE \cite{thoreau2022p} grounds part of the latent space to
physical layers, improving extrapolation on remote-sensing imagery.
Normalizing field flows \cite{guo2022normalizing} learn an invertible
map from a Gaussian process to a target field by maximizing likelihood
at scattered locations, unifying forward and inverse stochastic PDE
solution. Diffusion-based priors have also been coupled to PDE
constraints: DiffusionPDE \cite{huang2406diffusionpde} learns the joint
distribution of solution and coefficient fields to jointly in-paint
observations and solve the inverse problem, and physics-informed
distillation \cite{zhang2025physics} enforces the PDE residual post-hoc
rather than at every diffusion step, which enables single-step generation.

In this work, we propose \textit{GenVoid}, a physics-informed generative model-based framework for identifying voids in complex two- and three-dimensional domains. The main contributions of this work are as follows:

\begin{itemize}
\item We introduce \textit{GenVoid}, a physics-informed generative model-based framework that identifies internal voids in arbitrary two- and three-dimensional domains using surface displacement measurements alone. The framework is demonstrated across a range of material behaviours, including linear elasticity, hyperelasticity and plasticity, highlighting its applicability to diverse classes of solid mechanical systems. \textit{GenVoid} is capable of detecting multiple voids in both two- and three-dimensional domains.
\item  \textit{GenVoid} explicitly accounts for measurement uncertainty and noise inherent in surface displacement data. Rather than providing a deterministic estimate alone, the framework produces probabilistic void-identification outputs, thereby quantifying uncertainty in the inferred internal void geometry and providing a measure of confidence in the reconstruction.
\item We validate \textit{GenVoid} using both high-fidelity synthetic data and experimentally measured displacement fields acquired during in-situ mechanical experiments, demonstrating its ability to infer internal voids from realistic surface measurements.
\item  We introduce an observability measure that quantifies the sensitivity of boundary displacement measurements to localized stiffness perturbations within the interior of a structure. Specifically, observability characterizes how strongly a defect at a given spatial location influences the boundary sensor responses under an ensemble of applied loads, thereby addressing whether hidden defects can be reliably detected irrespective of where they form. Using this framework, we systematically investigate how the number and spatial distribution of boundary measurements govern void-identification accuracy.
\end{itemize}
Figure~\ref{fig:exp_pipeline} illustrates the experimental validation of void reconstruction using GeoVoid. Boundary displacements measured by digital image correlation (DIC) \cite{peters1982digital} from a physical specimen containing arbitrarily shaped voids are directly input to GenVoid. The resulting posterior inference provides a spatially resolved probability map of the underlying void geometry.

This paper is organized as follows. Section~\ref{sec:problem} formulates the void-identification problem, comprising the forward elasticity model that maps a candidate void to boundary displacements and the inverse problem of interest, recovering the void from these measurements. Section~\ref{sec:method} introduces \textit{GenVoid}, a physics-informed generative model for probabilistic void reconstruction. We first describe the level-set representation of the void and hybrid parameterization, which combines a learned shape prior with a local radial-basis function, followed by the network architecture and its variational latent space. We then define the training loss and accuracy metrics and introduce an observability measure that characterizes, independently of any specific defect, where within a material a void can in principle be detected. Section~\ref{sec:results} presents the results. We first evaluate two- and three-dimensional identification across diverse material geometries and convex and nonconvex void shapes, and then determine the minimum sensor density required for reliable reconstruction. We next assess reconstruction robustness to displacement measurement uncertainty and, evaluate the method under hyperelastic and plastic responses to test its robustness beyond linear elasticity. To this end, we also conduct deformation experiments on two-dimensional non-convex voids and use real experimental data obtained from DIC  to identify the void shapes with GenVoid. Section~\ref{sec:conclusions} concludes the paper. The symbols and abbreviations used throughout this manuscript are defined in Appendix~\ref{sec:nomenclature}.

\section{Problem Statement for Void Identification}\label{sec:problem}
Let $\Omega_0 \subset \mathbb{R}^3$ be a bounded open reference domain with boundary $\partial \Omega_0 = \Gamma_u \cup \Gamma_t$ ($\Gamma_u \cap \Gamma_t = \emptyset$, $|\Gamma_u| > 0$). Let $\Omega_\omega \subset\subset \Omega_0$ denote an unknown interior void domain(s) with boundary $\Gamma_w = \partial\Omega_\omega$, defining the actual material domain $\Omega = \Omega_0 \setminus \Omega_{\omega}$.

\vspace{0.2cm}
\noindent \textbf{Forward Problem}: For a candidate void $\omega \in \mathcal{O}_{\mathrm{ad}}$, find the displacement field $\mathbf{u}: \Omega \to \mathbb{R}^3$ satisfying the boundary value problem:
\begin{equation}\label{eq:forward}
\begin{aligned}
\nabla_{\mathbf{X}} \cdot \mathbf{P}(\mathbf{u}) + \mathbf{b}_0 &= \mathbf{0} \quad &&\text{in } \Omega, \\
\mathbf{u} &= \mathbf{u}_D \quad &&\text{on } \Gamma_u, \\
\mathbf{P}(\mathbf{u}) \hat{\mathbf{n}} &= \mathbf{t}_N \quad &&\text{on } \Gamma_t, \\
\mathbf{P}(\mathbf{u}) \hat{\mathbf{n}}_w &= \mathbf{0} \quad &&\text{on } \Gamma_w,
\end{aligned}
\end{equation}
where $\hat{\mathbf{n}}$ is the outward unit normal vector on the outer Neumann boundary $\Gamma_t$, and $\hat{\mathbf{n}}_w$ is the unit normal vector pointing into the void across $\Gamma_w$. $\mathbf{P}(\mathbf{u})$ is the First Piola--Kirchhoff stress tensor governed by the material constitutive law:
\begin{itemize}
    \item \textbf{Linear Elasticity:} $\mathbf{P} \approx \boldsymbol{\sigma} = \mathbb{C} : \boldsymbol{\varepsilon}$, where $\boldsymbol{\varepsilon} = \frac{1}{2}\left(\nabla \mathbf{u} + (\nabla \mathbf{u})^T\right)$ is the linearized strain tensor.
    \item \textbf{Hyperelasticity:} $\mathbf{P} = \frac{\partial W(\mathbf{F})}{\partial \mathbf{F}}$, with deformation gradient $\mathbf{F} = \mathbf{I} + \nabla_{\mathbf{X}} \mathbf{u}$ and strain energy density $W(\mathbf{F})$.
    \item \textbf{Plasticity (Deformation Theory):} $\mathbf{P} \approx \boldsymbol{\sigma} = \mathbf{s} + p \mathbf{I}$, governed by the non-linear isotropic Ramberg--Osgood relation \cite{ramberg1943description}:
    \begin{equation}\label{eq:ramberg_osgood}
    e_{\mathrm{eq}} = \frac{q}{3G} + \tau \frac{q}{E} \left( \frac{q}{\sigma_0} \right)^{n-1}, \qquad \mathbf{s} = \frac{2q}{3 e_{\mathrm{eq}}} \mathbf{e}, \qquad p = \kappa \, \mathrm{tr}(\boldsymbol{\varepsilon}),
    \end{equation}
    where $\mathbf{s}$ and $\mathbf{e} = \boldsymbol{\varepsilon} - \frac{1}{3}\mathrm{tr}(\boldsymbol{\varepsilon})\mathbf{I}$ are the stress and strain deviators, $q = \sqrt{\frac{3}{2} \mathbf{s} : \mathbf{s}}$ is the von Mises equivalent stress, $p$ is the hydrostatic pressure, $E$ and $G$ are Young's and shear moduli, $\kappa$ is the bulk modulus, $\sigma_0$ is the reference yield stress, $n$ is the strain hardening exponent, and $\tau$ is the yield offset parameter.
\end{itemize}

\vspace{0.2cm}
\noindent \textbf{Physics-Informed Inverse Problem}:
Let $\mathbf{u}_i^{\mathrm{0}}$ denote displacement measurements recorded at spatial sensor locations $\mathbf{X}_i \in \Gamma_s \subseteq \partial \Omega_0$ ($i = 1, \dots, N_s$). Reconstructing the unknown void $\omega^*$ is formulated as finding the optimal domain $\omega \in \mathcal{O}_{\mathrm{ad}}$ that minimizes the loss function between the FEM-simulated displacement $\mathbf{u}_h(\mathbf{X}_i; \omega)$ and observed displacements:
\begin{equation}\label{eq:inverse_problem}
\omega^* = \min_{\omega \in \mathcal{O}_{\mathrm{ad}}} \mathcal{J}(\omega; \mathbf{u}_h(\omega)) := \frac{1}{2} \sum_{i=1}^{N_s} \left\| ~\underbrace{\mathbf{u}_h(\mathbf{X}_i; \omega)}_{\substack{\text{FEM} \\ \text{Displacement}}} ~- ~\mathbf{u}_i^{\mathrm{0}} ~\right\|^2_{\mathbb{R}^3},
\end{equation}
\begin{equation*}
\text{subject to} \quad \mathbf{R}(\mathbf{u}_h(\omega);\omega) = \mathbf{0},
\end{equation*}
where $\mathcal{O}_{\mathrm{ad}}$ is the set of admissible void topologies, and $\mathbf{R}(\mathbf{u}_h(\omega);\omega)=\mathbf{F}_{\mathrm{int}}(\mathbf{u}_h;\omega)-\mathbf{F}_{\mathrm{ext}}(\omega) \in \mathbb{R}^{n_{\mathrm{dof}}}$ represents the \textit{global nonlinear residual vector} of the discretized Finite Element system.
It is obtained by discretizing the governing equilibrium equation, with the constitutive law determining the stress response and the void geometry defining the computational domain.

\section{Methodology}\label{sec:method}
We introduce \textit{GenVoid}, a physics-informed generative model for inferring the number, location and morphology of interior voids in a material from measurements obtained solely at its outer boundary. GenVoid is a \textit{parametric level-set} method that combines three key components: a learned shape prior that provides an informed initial estimate while constraining the search space; a correction basis that recovers local geometric detail beyond the prior; and a continuation strategy that progressively sharpens the material interface. The geometry is represented throughout using radial basis functions (RBFs), allowing voids to merge, split or disappear naturally without explicit front tracking.
\begin{figure}[h!]
\centering
\includegraphics[scale=0.5,clip=true]{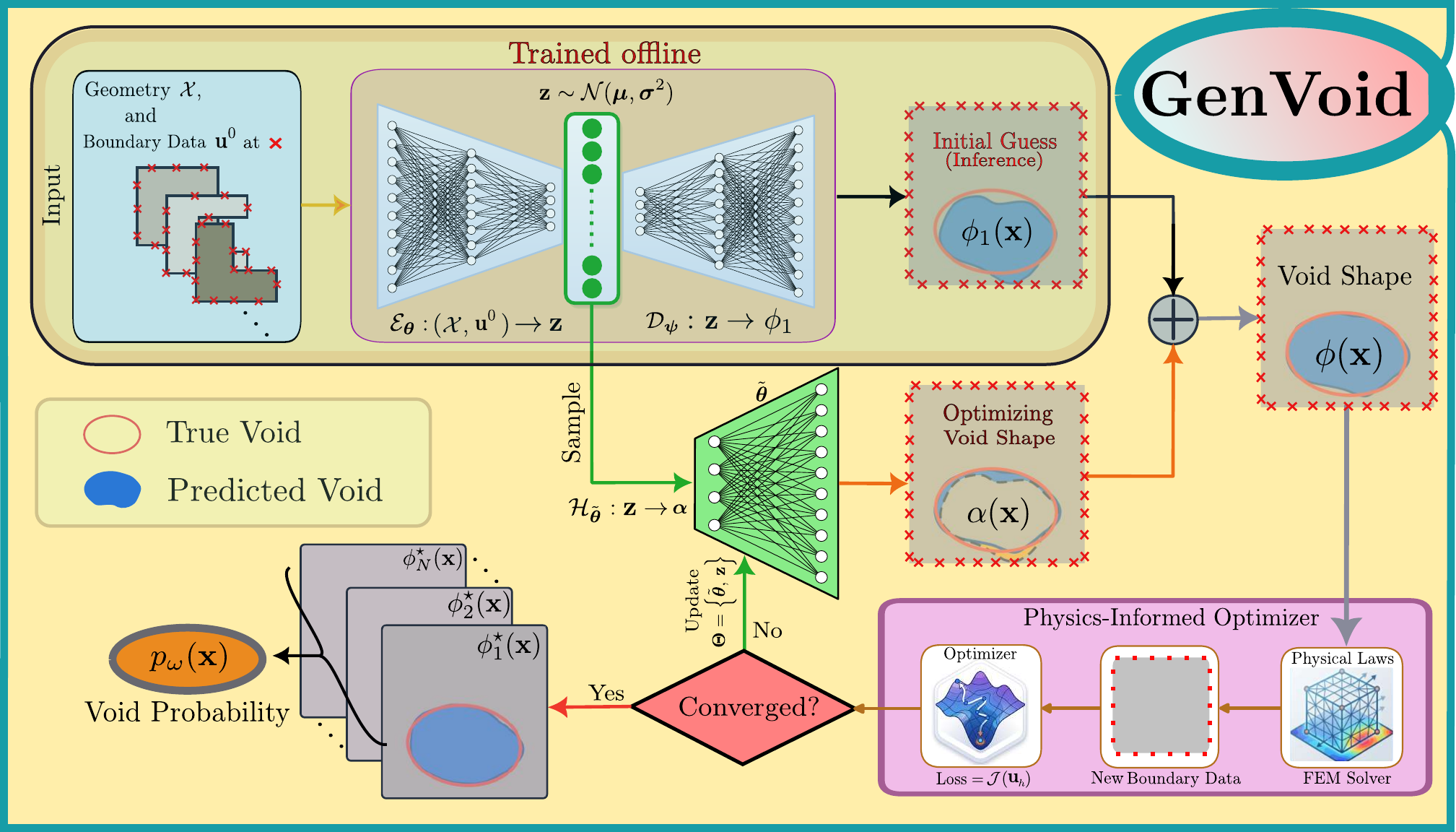}
\caption{\textbf{The GenVoid architecture}: Trained offline, the set encoder $\cE_{\boldsymbol{\theta}}$ maps the material geometry $\mathcal{X}$ and boundary measurements $\mathbf{u}^0$ to the parameters $\boldsymbol{\mu}$ and $\boldsymbol{\sigma}$ of a normal distribution from which $\mathbf{z}$ is sampled in the variational case. The decoder $\cD_{\boldsymbol{\psi}}$ transforms $\mathbf{z}$ into a prior level set, $\phi_1(\mathbf{x})$, providing the initial guess, while the MLP head $\cH_{\tilde{\boldsymbol{\theta}}}$ maps the same latent space to correction field  $\boldsymbol{\alpha}(\mathbf{x})$. Their sum defines the working level set, $\phi(\mathbf{x})$. The resulting geometry is passed to the physics-informed finite-element forward model, which solves the governing material physics and predicts the boundary response $\mathbf{u}_h$. The discrepancy between predicted and measured data is quantified by the loss function $\cJ(\mathbf{u}_h)$, and the optimizer updates $\boldsymbol{\Theta}=\{\tilde{\boldsymbol{\theta}},\mathbf{z}\}$ iteratively refine the solution till convergence. Repeated sampling from the latent distribution and optimization yields $\phi_1^*(\mathbf{x}),\ldots,\phi^*_N(\mathbf{x})$, from which the void probability at each spatial location, $p_{\omega}(\mathbf{x})$, is obtained.}
\label{fig:method}
\end{figure}
Figure~\ref{fig:method} shows the schematic representation of GenVoid for the void identification framework.

\subsection{Level-Set Representation}
\label{sec:levelset}
We represent the interior void geometry using a parametric level-set function. The void region is defined by
$\Omega_{\omega} = \left\{ \mathbf{x} \in \Omega_0 : \phi(\mathbf{x}) < 0 \right\},$
where $\phi(\mathbf{x})$ is the level-set function represented as
\begin{equation}
\label{eq:rbf}
\phi(\mathbf{x}) = \sum_{j=1}^{N_R} a_j g_j(\mathbf{x}),
\end{equation}
where
\begin{equation}
\label{eq:rbf_kernel}
g_j(\mathbf{x}) = \left[\max(0,1-r_j)\right]^4(4r_j+1), \qquad r_j = \sqrt{\delta \left|\mathbf{x}-\mathbf{x}_j\right|^2+c^2}.
\end{equation}
Here, $\mathbf{x}_j$ denotes the center of the $j$th RBF, $\mathbf{a}=[a_1,\ldots,a_{N_R}]^{\text{T}}$ is the coefficient vector, and
$N_R = N_{kx} N_{ky} N_{kz}$
is the total number of RBF centers on the regular
$N_{kx}\times N_{ky}\times N_{kz}$ knot grid. The kernel in Eq.~\eqref{eq:rbf_kernel} is the compactly supported Wendland $C^2$ kernel \cite{wendland1995piecewise}. The parameter $\delta$ controls its spatial support, which is chosen to give a support radius of $2.2$ knot spacings. We use $c=10^{-3}$. The geometry is encoded entirely by $\mathbf{a}$. Consequently, a single coefficient vector can represent multiple disconnected voids with different shapes and can accommodate topology changes, including the merging, splitting, and disappearance of voids, without explicit front tracking or remeshing.

\vspace{0.2cm}
\noindent \textbf{Smoothed Material Interpolation}:
The level-set function is mapped to the finite-element material properties through an \textit{Ersatz-material interpolation} \cite{allaire2004structural}. At each finite element node $\mathbf{x}_i$, the level-set value is
\begin{equation}
\label{eq:phi_nodes}
\phi(\mathbf{x}_i) = \sum_{j=1}^{N_R} \mathcal{G}_{ij}a_j, \qquad \mathcal{G}_{ij}=g_j(\mathbf{x}_i),
\end{equation}
where $\mathcal{G}$ is the sparse \textit{node-to-knot} matrix. The effective Young's modulus of element $e$ is defined as
\begin{equation}
\label{eq:ersatz}
E_e(\mathbf{a}) = \frac{E_0}{8} \sum_{i \in e} H_b\left(\phi(\mathbf{x}_i)\right)^2,
\end{equation}
where the sum is taken over the eight nodes of hexahedral element $e$ (in three-dimensions) and $E_0$ denotes the Young's modulus of the surrounding material. The square of the smoothed Heaviside function is used as the Ersatz-material interpolation. Therefore, a region fully contained within a void has an effective modulus of approximately
$E_{\mathrm{void}} \approx \gamma^2 E_0,$
where $\gamma=10^{-3}$ is the stiffness floor.
The smoothed Heaviside function is defined as
\begin{equation}
\label{eq:heaviside}
H_b(\phi) = 
\begin{cases}
\gamma, & \phi < -b, \\[4pt]
\displaystyle \frac{3(1-\gamma)}{4} \left( \frac{\phi}{b} - \frac{\phi^3}{3b^3} \right) + \frac{1+\gamma}{2}, & |\phi| \le b, \\[8pt]
1, & \phi > b.
\end{cases}
\end{equation}
Its derivative is
\begin{equation}
\label{eq:heaviside_derivative}
H_b'(\phi) = 
\begin{cases}
\displaystyle \frac{3(1-\gamma)}{4b} \left( 1 - \frac{\phi^2}{b^2} \right), & |\phi| \le b, \\[8pt]
0, & |\phi| > b.
\end{cases}
\end{equation}
The interface width $b$ is a numerical regularization parameter with no physical interpretation. It smooths the transition between material and void and ensures differentiability of the element stiffness with respect to $\mathbf{a}$. Its nominal value is
$b_0 = 2.5h,$
where $h$ denotes the smallest element size.

\vspace{0.2cm}
\noindent \textbf{Finite Element Forward Solver}:
The effective element stiffnesses are used to assemble the global stiffness matrix,
\begin{equation}
\label{eq:global_stiffness}
\mathbf{K}(\mathbf{a}) = \sum_e E_e(\mathbf{a}) \mathbf{K}_0^{(e)},
\end{equation}
where $\mathbf{K}_0^{(e)}$ denotes the assembled contribution of the reference element stiffness matrix. Since all elements use the same reference stiffness matrix, it is constructed only once.
For the linear problem, the forward system is solved using sparse LU factorization. For a nonlinear constitutive law, the linear solve is replaced by a Newton iteration for each load case, as described in Section~\ref{sec:nonlinear}.

\subsection{Neural Network Architecture}
\label{sec:arch}
The RBF representation of the level-set field has far more learnable parameters than the boundary measurements can meaningfully constrain. Direct optimization of RBF coefficients is therefore poorly conditioned and can depend strongly on the initial guess. Moreover, $H_b'(\phi_i)$ vanishes for $|\phi_i|>b$, so the gradient can modify an existing boundary but cannot act on material outside the active region. A void placed far from its correct location may consequently receive little or no gradient indicating where it should move.
The space of admissible voids is considerably more structured than the full coefficient space $\mathbb{R}^{N_R}$. 
We therefore introduce a lower-dimensional latent representation of the admissible geometries and perform the subsequent identification in this latent space. A geometry is represented by a latent variable $\mathbf{z}$ rather than by $N_R$ independent coefficients, and a decoder maps $\mathbf{z}$ to the RBF coefficients that define the level-set field.

\subsubsection{Variational Latent Representation}
\label{sec:vae}

To capture geometrical features that are only weakly constrained by boundary observations, we adopt a variational latent representation. Rather than collapsing each shape into a single deterministic point, we model its latent state as a Gaussian distribution parameterized by a mean $\boldsymbol{\mu}$ and standard deviation $\boldsymbol{\sigma}$. Stochastic samples are drawn via the reparameterization
\begin{equation}
\label{eq:vae}
\mathbf{z} = \boldsymbol{\mu} + \boldsymbol{\sigma} \odot \boldsymbol{\epsilon},
\qquad
\boldsymbol{\epsilon} \sim \mathcal{N}(\mathbf{0}, \mathbf{I}),
\end{equation}
ensuring gradients propagate smoothly through the probabilistic bottleneck. A Kullback--Leibler (KL) divergence penalty subsequently constrains these distributions toward a standard normal prior, establishing a unified coordinate scale and preventing unbounded growth in latent variance.

The decoder and encoder are trained in two separate stages, each responsible for a distinct task. In the first stage, the decoder learns to represent void geometry: given a latent vector, it produces the corresponding level-set field. Following an auto-decoder strategy, each training shape is assigned learnable latent parameters $(\boldsymbol{\mu}_k, \boldsymbol{\sigma}_k)$ that are optimized directly alongside the decoder weights against the known target field $\phi^{(k)}$, without involving an encoder. This ensures the latent space is shaped purely by the geometry of the training data. In the second stage, the decoder weights are held frozen while the encoder is trained to map boundary measurements to latent vectors within the pre-established space. This decoupling prevents the encoder's initial optimization difficulty from distorting the latent topology. Furthermore, training the encoder \emph{through} the frozen decoder via the field reconstruction loss, rather than regressing directly against the Stage 1 latent vectors, avoids issues of latent coordinate non-uniqueness arising from symmetries or transformations.

\vspace{0.2cm}
\noindent \textbf{Decoder Training}:
\label{sec:decoder}
The model is trained in two stages. First, the decoder is trained together with a latent distribution associated with each training geometry. The training data contains configurations comprising spheres, ellipsoids, lobed shapes, tori, crescents, L-shaped and cross-shaped voids, as well as random combinations of two or three such primitives with varying scale and pose. For each configuration $k$, the corresponding level-set field $\phi^{(k)}$ is used as the reconstruction target. The geometry is assigned learnable latent parameters $(\boldsymbol{\mu}_k,\boldsymbol{\sigma}_k)$, from which $\mathbf{z}_k$ is sampled according to Eq.~\eqref{eq:vae}. The decoder parameters and latent distributions are optimized jointly according to
\begin{equation}
\label{eq:vae_loss}
\min_{\boldsymbol{\psi}} \; \mathcal{L}_{\mathrm{dec}} = \sum_k \left[ \left\| \mathcal{D}_{\boldsymbol{\psi}}(\mathbf{z}_k) - \phi^{(k)} \right\|_2^2 + \beta D_{\mathrm{KL}}\left( q_k(\mathbf{z}) \parallel \mathcal{N}(\mathbf{0},\mathbf{I}) \right) \right],
\end{equation}
where $q_k(\mathbf{z})$ denotes the Gaussian latent distribution for geometry $k$ and $\beta$ controls the strength of the KL regularization. No encoder is required at this stage: since the exact target field $\phi^{(k)}$ is already known for every training shape, a suitable latent distribution can be found by direct optimization rather than by inference, so $(\boldsymbol{\mu}_k,\boldsymbol{\sigma}_k)$ is optimized jointly with the decoder weights as a free parameter, and the resulting latent space is shaped purely by the geometry of the training data rather than by any measurement process. The decoder $\mathcal{D}_{\boldsymbol{\psi}}$ maps the latent variable directly to the RBF coefficient vector. It comprises four fully connected layers with widths $512$, $1{,}024$, $1{,}024$ and $N_R$. The first three layers are followed by Layer Normalization and SiLU activation. A separate volumetric generator is unnecessary because the RBF lattice in Eq.~\eqref{eq:rbf} already provides a compact representation of the level-set field. Predicting $N_R=1{,}024$ coefficients is therefore substantially more economical than predicting a dense three-dimensional field.
For a latent sample $\mathbf{z}$, the decoded coefficients define the level-set field:
\begin{equation}
\label{eq:decoded_phi}
\phi_1 = \mathcal{G} \mathcal{D}_{\boldsymbol{\psi}}(\mathbf{z})
\end{equation}

\vspace{0.2cm}
\noindent \textbf{Encoder Training}:
\label{sec:encoder}
In the second stage, the decoder $\mathcal{D}_{\boldsymbol{\psi}}$ is frozen and an encoder is trained to infer the latent distribution from the boundary observations and domain geometry. The training set contains $13{,}008$ simulated pairs $(\mathbf{u}_k^0,\phi^{(k)})$, with one to four voids randomly distributed within the material domain.
The encoder $\mathcal{E}_{\boldsymbol{\theta}}$ predicts the mean and standard deviation of the latent distribution,
\begin{equation}
\label{eq:encoder_output}
(\boldsymbol{\mu}_k,\boldsymbol{\sigma}_k) = \mathcal{E}_{\boldsymbol{\theta}}\left( \mathcal{X}_k, \mathbf{u}_k^0 \right),
\end{equation}
and a latent sample is then obtained from Eq.~\eqref{eq:vae}. The encoder is trained through the frozen decoder by minimizing
\begin{equation}
\label{eq:encoder_loss}
\min_{\boldsymbol{\theta}} \; \mathcal{L}_{\mathrm{enc}} = \sum_k \left[ \left\| \mathcal{D}_{\boldsymbol{\psi}}(\mathbf{z}_k(\boldsymbol{\theta})) - \phi^{(k)} \right\|_2^2 + \beta D_{\mathrm{KL}}\left( q_k(\mathbf{z};\boldsymbol{\theta}) \parallel \mathcal{N}(\mathbf{0},\mathbf{I}) \right) \right].
\end{equation}

The decoder is kept frozen in this stage so that the latent space learned in Stage~1 is not disturbed by measurement noise or by the encoder's own optimization: only $\mathcal{E}_{\boldsymbol{\theta}}$ is updated, with gradients passing through the fixed decoder to reach it. This formulation also avoids regressing the observations against the latent coordinates obtained during decoder pretraining, since such coordinates are not unique, because rotations and other transformations of the latent space can leave the reconstructed geometries unchanged; instead, the encoder is trained directly through the geometric reconstruction objective.

\paragraph{Remark 1:} The two-stage strategy ensures that the latent space is optimized purely for geometric representation before training the encoder to map indirect boundary measurements into it. Simultaneous training would otherwise create a destructive optimization tug-of-war between learning shape features and solving an ill-posed inverse problem. Isolating the decoder in the first stage allows it to forge a pristine, uncorrupted manifold using exact, full-field targets without encoder interference. Freezing the decoder in the second stage locks down that geometric atlas as a stable, stationary coordinate system for the remainder of training. This deliberate decoupling prevents the encoder's initial struggles from degrading the quality of the representation space. 

\vspace{0.2cm}
\noindent \textbf{Permutation-Invariant Encoding of Boundary Measurements}:
\label{sec:set_encoder}
To accommodate varying boundary discretizations and measurement configurations, the encoder inputs both domain geometry coordinates $\mathcal{X}$ and boundary displacement measurements $\mathbf{u}^0$ as an unordered set. This formulation ensures invariance to fluctuations in station counts and arbitrary index permutations. For each measurement station $s$, we define a composite feature vector $\boldsymbol{\xi}_s \in \mathbb{R}^{3+3N_l}$ comprising its spatial coordinates $\mathbf{X}_s$ alongside displacement observations across $N_l$ distinct load cases:
\begin{equation} \label{eq:station_feature}
\boldsymbol{\xi}_s = \begin{bmatrix} \mathbf{X}_s & \mathbf{u}_s^{0,1} & \cdots & \mathbf{u}_s^{0,N_l} \end{bmatrix}^T
\end{equation}
To ensure permutation invariance and robustness to arbitrary station counts and orderings, the encoder aggregates the station features $\boldsymbol{\xi}_s$ via symmetric global pooling. Concatenating the element-wise maximum and mean across all stations yields a fixed-length representation that is independent of station indexing, from which the latent distribution parameters $(\boldsymbol{\mu}, \boldsymbol{\sigma})$ are evaluated:
\begin{equation}
\label{eq:setenc}
\mathcal{E}_{\boldsymbol{\theta}}\left( \mathcal{X}, \mathbf{u}^0 \right)
=
\mathcal{E}_{\boldsymbol{\theta}}\Big(
\big[
\operatorname{max}_s \boldsymbol{\xi}_s,\,
\operatorname{mean}_s \boldsymbol{\xi}_s
\big]
\Big).
\end{equation}

\paragraph{Remark 2:} While formulated above for three-dimensional domains, this architecture adapts directly to two dimensions by redefining only the spatial discretization and network shapes; equations \eqref{eq:rbf}, \eqref{eq:ersatz}, and \eqref{eq:heaviside} hold for $\mathbf{x}\in\R^2$, and the Wendland kernel maintains its $C^2$ continuity in dimensions up to three. 
 Measurement stations are positioned along boundary surfaces, with each station capturing two displacement components per load case such that $\boldsymbol{\xi}_s\in\R^{2+2N_l}$ in \eqref{eq:setenc}.

\subsubsection{Hybrid Level-Set Representation}
Because a learned parameterization alone cannot capture arbitrary fine geometry, whereas an unconstrained RBF expansion suffers from poor conditioning, we model the total level-set field as a hybrid sum (Fig.~\ref{fig:method}):
\begin{equation}
\label{eq:model}
\phi(\mathbf{z}) = \underbrace{\mathcal{G} \mathcal{D}_{\boldsymbol{\psi}}(\mathbf{z})}_{\phi_1(\mathbf{x})} + \underbrace{\mathcal{G} \mathcal{H}_{\tilde{\boldsymbol{\theta}}}(\mathbf{z})}_{\boldsymbol{\alpha}(\mathbf{x})}
\end{equation}

where $\mathcal{G}$ is the node-to-knot interpolation matrix from Eq.~\eqref{eq:rbf}. Here, the decoder $\mathcal{D}_{\boldsymbol{\psi}}$ provides the primary structural topology, specifying void count and approximate location, while the RBF correction term (MLP head) supplies localized boundary refinement beyond the decoder’s expressive limit. Importantly, the correction field $\boldsymbol{\alpha}(\mathbf{x})$ serves solely as a boundary perturbation rather than an independent geometry: its zero level set lacks physical meaning, and its amplitude is restricted to an order of magnitude below $\phi_1(\mathbf{z})$ via regularization. Evaluating the summation in Eq.~\eqref{eq:model} prior to applying the sign operator prevents simple geometric superposition. Far from boundaries, where $|\phi_1|$ is large, the correction term is negligible; thus, it operates exclusively near interfaces ($\phi_1 \approx 0$) to fine-tune wall positions.
The correction network $\mathcal{H}_{\tilde{\boldsymbol{\theta}}}$ comprises three fully connected layers of width $512$ with Layer Normalization and SiLU activations. Its output layer is initialized to \textit{zero}, guaranteeing that $\boldsymbol{\alpha} = \mathbf{0}$ and $\phi = \phi_1$ at initialization. Unlike the decoder, the MLP parameters $\tilde{\boldsymbol{\theta}}$ are not pretrained on a data; instead, they are treated as site-specific unknowns optimized jointly with $\mathbf{z}$ during inverse reconstruction.

\subsubsection{Probabilistic Void Identification via Monte Carlo Sampling}
During training, latent vectors are sampled from the variational Gaussian distributions via the reparameterization trick, while a KL-divergence penalty anchors the latent space to a standard normal prior. This yields a continuous latent manifold rather than isolated point estimates, ensuring that neighboring latent samples decode into physically plausible geometries.

To quantify geometric uncertainty under noisy boundary observations, we employ a \textit{Monte Carlo} approach that evaluates the latent posterior distribution conditioned on the measured data. To account for observation-specific shifts, the downstream MLP head is first fine-tuned for the given realization and subsequently held fixed. We then draw $N >>1$ latent samples $\{\mathbf{z}_i\}_{i=1}^{N}$ from the posterior, passing them through the frozen decoder and adapted MLP head to generate an ensemble of candidate level-set fields $\{\phi_i^*(\mathbf{x})\}_{i=1}^{N}$. The spatial probability of a domain point $\mathbf{x}$ residing within a void is subsequently estimated via this Monte Carlo ensemble as the empirical fraction of realizations returning a negative level-set value: \begin{equation} \label{eq:pvoid} p_{\omega}(\mathbf{x}) = \frac{1}{N} \sum_{i=1}^{N} \mathbf{1}\left[\phi_i^*(\mathbf{x}) < 0\right]. \end{equation} This formulation translates the sampled ensemble into a spatially resolved probability map, where $p_{\omega}(\mathbf{x}) = 1$ denotes regions unanimously classified as void across all realizations, and $p_{\omega}(\mathbf{x}) = 0$ indicates entirely solid domains.

\subsection{Metrics and Loss Function}\label{sec:loss}
To quantify reconstruction fidelity, we define a probabilistic soft intersection-over-union ($\mathrm{IoU}_p$). Let the predicted void be represented by a probability field $p_{\omega}(x)\in[0,1]$ over the entire domain $\Omega_0$, and let $y(x)=\mathbf{1}_{\omega^{\rm true}}(x)$ denote the indicator function of the ground-truth void region. The continuous soft $\mathrm{IoU}_p$ is defined as
\begin{equation}
\label{eq:softiou1}
\mathrm{IoU}_p=
\frac{
\int_{\Omega} p_{\omega}(x)y(x)~\mathrm{d}x
}{
\int_{\Omega}
\left[p_{\omega}(x)+y(x)-p_{\omega}(x)y(x)\right]~\mathrm{d}x
}
\in[0,1],
\end{equation}
where the numerator and denominator correspond to the soft intersection and soft union, respectively.
On a voxelized grid comprising $N$ elements, the corresponding discrete form is
\begin{equation}
\label{eq:softiou2}
\mathrm{IoU}_p=
\frac{
\sum_{i=1}^{N} p_{{\omega}_i} y_i
}{
\sum_{i=1}^{N}
\left(p_{{\omega}_i}+y_i-p_{{\omega}_i} y_i\right)
},
\qquad
y_i=\mathbf{1}_{\omega^{\rm true}}(x_i),
\end{equation}
where the predicted occupancy is weighted by $p_i$, such that voxels assigned high probability contribute more strongly to the overlap measure than those with marginal occupancy probabilities.

For a deterministic realization of the latent representation, for which $p_{\omega}(x)\in \{0,1\}$ everywhere, Eq.~\eqref{eq:softiou1} reduces to the following volumetric overlap,
\begin{equation}
\label{eq:iou}
\mathrm{IoU}
=
\frac{
\left|\omega^{\rm pred}\cap\omega^{\rm true}\right|
}{
\left|\omega^{\rm pred}\cup\omega^{\rm true}\right|
}
\in[0,1].
\end{equation}

The loss function $\mathcal{J}(\boldsymbol{\Theta})$ for the inverse void identification problem, defined over $N_l$ load cases and $N_s$ measurement stations as given as follows:
\begin{equation}
\label{eq:loss}
\mathcal{J}(\boldsymbol{\Theta})=\sum_{l=1}^{N_l}\sum_{s=1}^{N_s} \left\lVert\mathbf{u}_{h,s}^{l}(\boldsymbol{\Theta})-\mathbf{u}_s^{0,l}\right\rVert_2^2 +  \lambda\left\lVert\tilde{\boldsymbol{\theta}}\right\rVert_2^2,
\end{equation}
where $\boldsymbol{\Theta} = \{\boldsymbol{\tilde{\theta}}, \mathbf{z}\}$ are the trainable parameters, $\mathbf{u}_{h,s}^{l}(\boldsymbol{\Theta})$ is the displacement predicted at station $\mathbf{X}_s$ under load case $l$ and $\mathbf{u}_s^{0,l}$ the corresponding measurement. The second term is the \textit{Tikhonov regularization} term for MLP head weights.
The loss reaches the unknown void only through the coefficients $\mathbf{a}\in\R^{N_R}$ of \eqref{eq:rbf}, which fix the nodal level set $\phi_i=\sum_j \mathcal{G}_{ij}a_j$, and through it the element moduli $E_e$ of \eqref{eq:ersatz}, the stiffness $\mathbf{K}$ and the predicted boundary displacements.

\vspace{0.2cm}
\noindent \textbf{Physics-Informed Generative Model}: What makes the identification physics-informed is that $\mathbf{u}_h$ is never approximated by a neural network; instead, it is obtained from the forward solver for the current candidate geometry. Each evaluation of the loss therefore entails a genuine finite-element solver, and the subsequent identification steps are differentiated through this same governing physics rather than through a learned surrogate. We employ FEniCS finite-element solver \cite{logg2012automated} to formulate and solve the governing equations on the structured mesh described in Section~\ref{sec:levelset}, for all three material models considered here. The linear-elastic model requires a single linear solve, whereas the hyperelastic and plastic models described in Section~\ref{sec:nonlinear} are solved iteratively using Newton--Raphson iterations. The same framework is used to compute the adjoint sensitivities. 

 \paragraph{Remark 3:} The loss function $\mathcal{J}(\boldsymbol{\Theta})$ is minimized using the Levenberg--Marquardt (LM) algorithm \cite{levenberg1944method, marquardt1963algorithm}. This makes it well suited to the identification problem here, where every step requires a finite-element solver, so a reliable descent direction is essential. The search is initialized from the encoder's prediction, $\mathcal{E}_{\boldsymbol{\theta}}(\mathcal{X},\mathbf{u}^0)$, with the MLP head initialized to zero so that $\phi=\phi_1$ at the first iteration, and proceeds until the loss falls below a tolerance value.

 \paragraph{Remark 4:} Three-dimensional domains are discretized using eight-node hexahedral elements, while two-dimensional domains are discretized using four-node quadrilateral elements. All other aspects of the formulation remain unchanged between the two cases.

\subsection{Spatial Observability and Identifiability of Hidden Defects}
\label{sec:obs}
\textit{Can boundary sensors reliably detect a hidden interior voids regardless of where it forms within a structure?} This is an important question to ask when assessing the fundamental limits of boundary displacement-based inverse void identification problem. In non-destructive evaluation and structural health monitoring, inverse problem solvers reconstruct hidden voids or cracks using measurements taken along the outer surface. However, defects situated far from sensors or within low-stress regions induce minimal perturbation in surface displacements. To quantify these spatial limits prior to optimization, we formalize this dependence through an observability map, $O(\mathbf{x})$, which evaluates whether a localized stiffness perturbation at an arbitrary interior location $\mathbf{x}$ produces a detectable displacement signature at the boundary sensors.

Let $E(\mathbf{x})$ denote the material Young's modulus at an interior position $\mathbf{x}$, where introducing a localized void reduces $E(\mathbf{x})$. For load case $l$, letting $\mathbf{u}_{h,l}$ represent the predicted boundary measurement vector and $\partial\mathbf{u}_{h,l}/\partial E(\mathbf{x})$ its sensitivity to localized stiffness variations, the scalar observability field is defined as the root-sum-square sensitivity across all $N_l$ load cases:
\begin{equation}
\label{eq:obs}
O(\mathbf{x}) = \left(\sum_{l=1}^{N_l} \left\lVert\frac{\partial\mathbf{u}_{h,l}}{\partial E(\mathbf{x})}\right\rVert_2^2\right)^{1/2}
\end{equation}
Physically, $O(\mathbf{x})$ quantifies how strongly a localized stiffness perturbation at $\mathbf{x}$ propagates to boundary sensor locations under the applied load ensemble. Formulated as a topological sensitivity, $O(\mathbf{x})$ is computed via adjoint solves that reuse the factorized stiffness matrix, requiring only one back-substitution per load case to evaluate the entire domain. Because $O(\mathbf{x})$ depends solely on domain geometry, load configurations, and sensor locations, remaining entirely independent of defect geometry; it can be evaluated prior to solving the inverse problem.

Observability peaks along free surfaces characterized by elevated stress and sensor proximity, decaying rapidly into the interior and toward clamped boundaries. Every load case is a unit point force: the cases differ in direction and in point of application. Directions alternate between the two orthogonal axes and the application point along the free edges, so each case is a distinct load pattern that adds observability rank rather than a rescaling of the previous one. Under a single static load ($N_l=1$), directional blind spots emerge as null lines where local strain energy density vanishes relative to measurement degrees of freedom. Applying multi-load excitation ensembles rotates these null axes and elevates the baseline domain sensitivity; increasing the ensemble to $N_l=10$ load cases increases the minimum value of $O(\mathbf{x})$ domain-wide by approximately an order of magnitude. Multi-load excitation thus eliminates localized directional blind spots and substantially mitigates ill-posedness throughout the domain, motivating our choice of $N_l=10$.

\begin{figure}
\centering
\includegraphics[width=\textwidth]{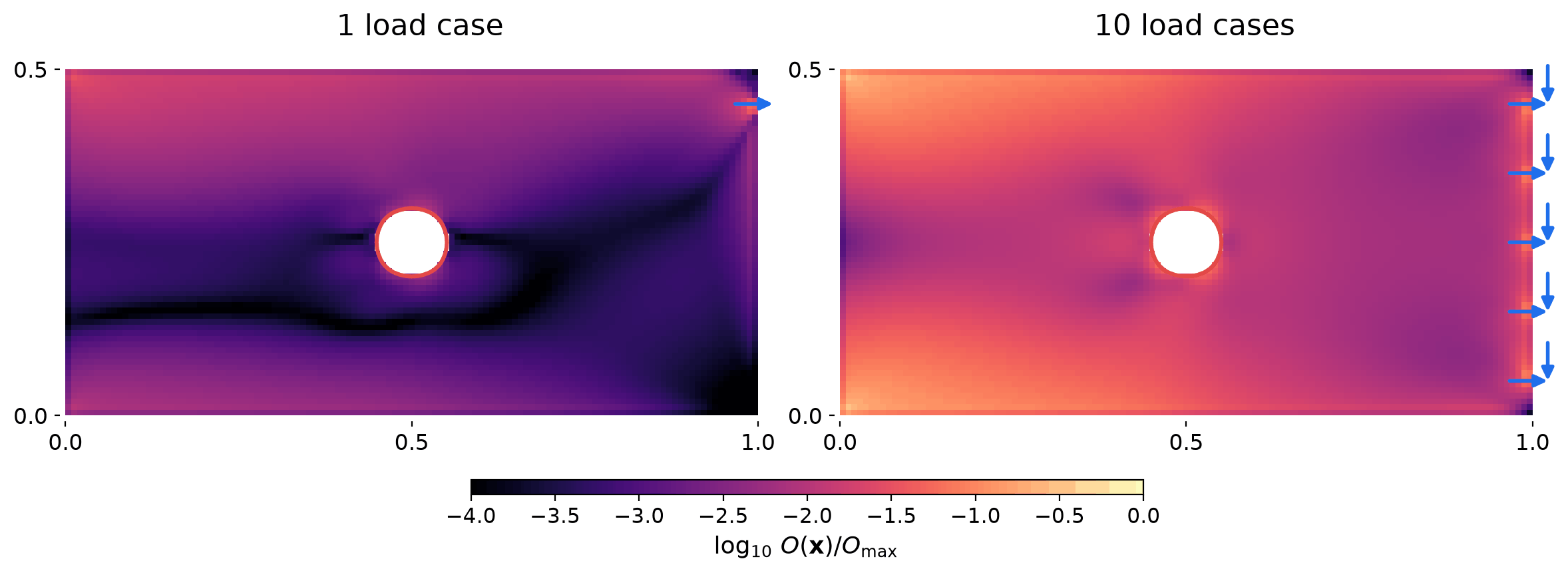}
\caption{Displacement observability on a domain with an existing circular
void (red outline), for one load case (left) and ten (right). The load set is
drawn as blue arrows: five stations spaced uniformly from top to bottom down
the free edge, each carrying two orthogonal unit loads, one normal to the edge
and one tangential to it, giving ten cases. Colour encodes
$\log_{10}\big(O(\mathbf{x})/O_{\max}\big)$ with $O_{\max}$ taken over both panels,
so brightness is directly comparable between them. The
void's interior is masked in white: $O(\mathbf{x})$.}
\label{fig:obsvoid}
\end{figure}

\vspace{0.2cm}
\noindent \textbf{Observability in the Presence of Existing Defects}:
The initial observability map is computed for a homogeneous domain; introducing a structural defect alters the background stress state against which subsequent sensitivity is evaluated. Figure~\ref{fig:obsvoid} therefore re-evaluates the displacement observability field in the presence of a circular void centred at $(0.5, 0.25)$, considering both a single load case and an ensemble of ten load cases. Both panels share a common logarithmic magnitude scale, normalised by the larger of the two peak values, enabling direct comparison of observability across the two loading conditions. Increasing the number of load cases enhances observability throughout the material, as shown in Figure~\ref{fig:obsvoid}.

Two low-sensitivity regions emerge immediately adjacent to the void along the principal loading trajectory. These regions represent defect-induced \emph{stress shadows}, arising because the existing void redirects internal force paths around the defect and thereby reduces the local strain-energy density. As a result, subsequent material perturbations within these regions produce only weak changes in the boundary displacement field.
Importantly, these stress shadows are governed by the presence of the defect itself rather than by the choice of loading configuration. Increasing the number of load cases therefore improves the overall observability of the domain but does not fully eliminate these localized regions of reduced sensitivity. This has a direct implication for sequential inspection: a second void located close to an existing defect may be substantially more difficult to detect than an isolated defect, creating a potential failure mode in which two neighbouring defects are interpreted as a single anomaly.
The interior of the existing void is excluded from the observability map, as topological sensitivity is not well defined within a region that has already been identified as void.

\section{Results}\label{sec:results}
This section systematically evaluates the performance of GenVoid in reconstructing complex void shapes across diverse geometries and material systems. Section~\ref{sec:2d3d} evaluates
void identification accuracy across a diverse set of geometries as well as void
shapes, including both convex and non-convex cases in two and three
dimensions. Section~\ref{sec:rbfcomp} compares this performance against a
reference solver that optimizes the RBF coefficients directly, without a
learned shape prior. Section~\ref{sec:sensors} examines the dependence of
void identification accuracy on the number of boundary measurement stations,
and Section~\ref{sec:uq} evaluates robustness to measurement noise (under uncertainty). Section~\ref{sec:nonlinear} extends
the evaluation beyond linear elasticity to hyperelastic and plastic
material models, to assess whether the identification accuracy is
maintained under more general constitutive behavior. To this end, in Section~\ref{sec:EXP}, we employ GenVoid to learn the two-dimensional voids for linear and nonlinear materials for which physical experiments are conducted and corresponding experimental displacement measurement data are obtained.

\subsection{Two- and Three-dimensional Void Identification}
\label{sec:2d3d}

\begin{figure}
\centering
\includegraphics[width=0.9\textwidth]{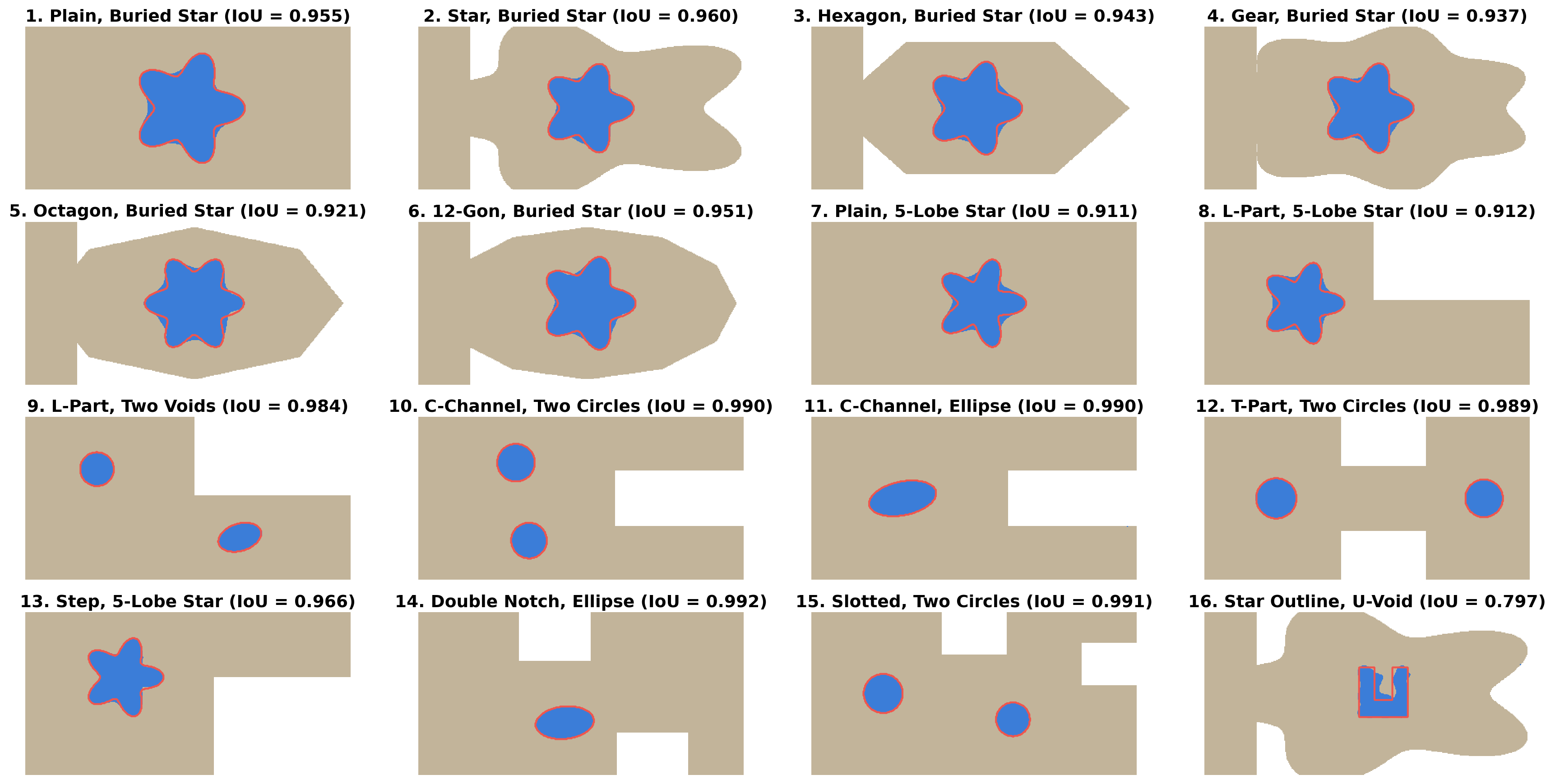}
\includegraphics[width=0.55\textwidth]{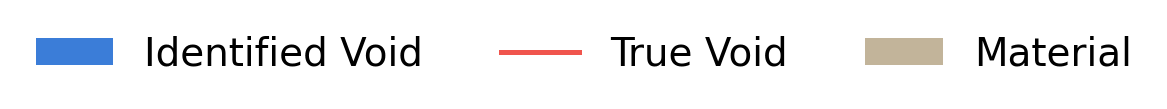}
\caption{Two-dimensional identification across sixteen different geometries containing voids of unknown shapes. Blue is the identified void, the red line is the true one, and the surrounding shaded region is the material. 
}
\label{fig:gallery2d}
\end{figure}

\begin{figure}
\centering
\includegraphics[width=0.9\textwidth]{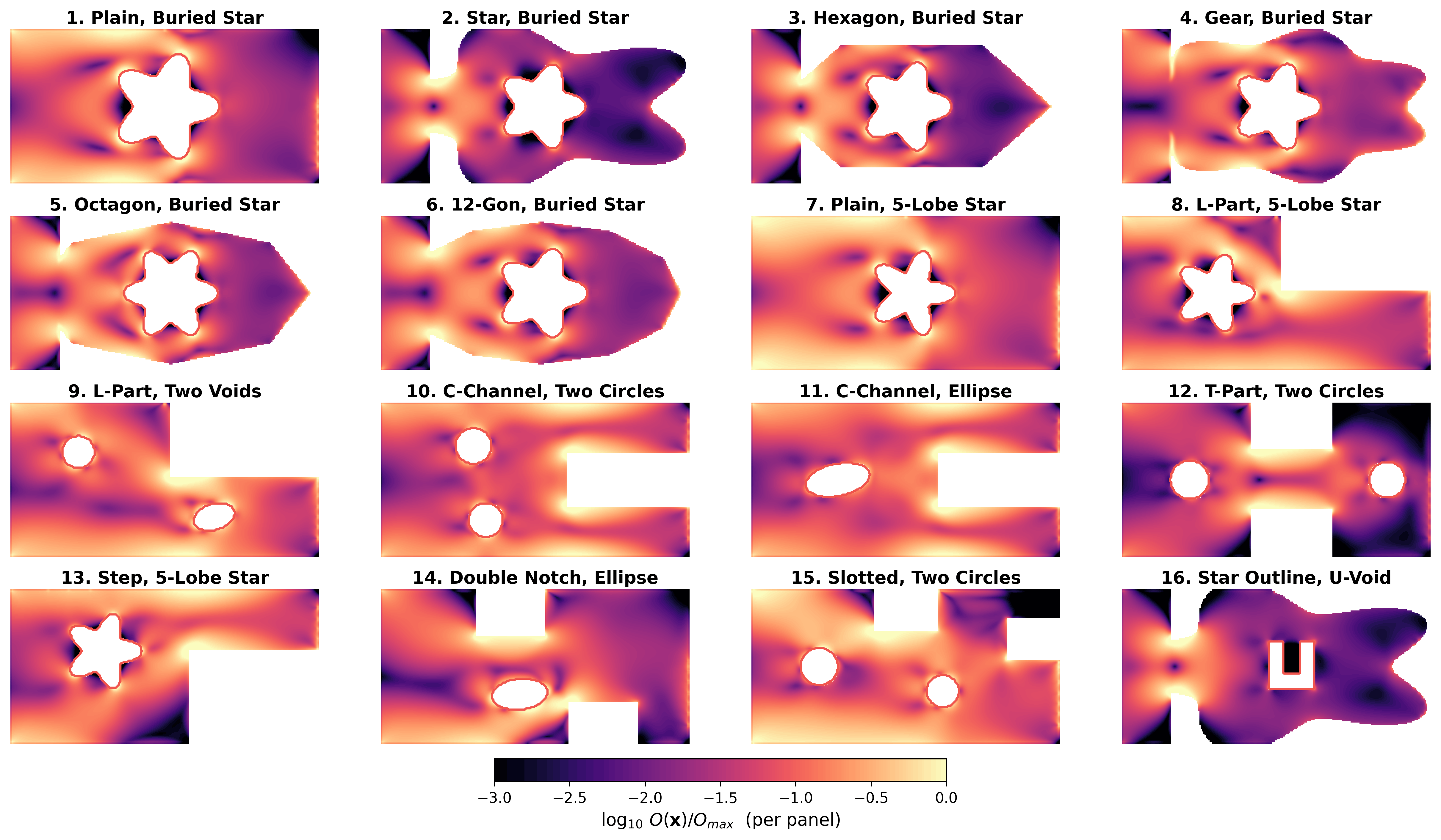}
\caption{Observability field $O(\mathbf{x})$ for the 16 material–geometry
cases of Figure~\ref{fig:gallery2d}, evaluated in the presence of each
case's true void under the $N_l=10$ applied loads.}
\label{fig:2d_obs}
\end{figure}

We systematically evaluated GenVoid across sixteen distinct two-dimensional geometries containing unknown internal voids; see, Fig.~\ref{fig:gallery2d}, encompassing both canonical shapes (rectangles, hexagons and octagons) and different complex geometries (gears, L-parts, C-channels, T-parts, steps and double notches). Void reconstruction accuracy, quantified by the IoU, ranged from 0.911 for a five-lobe star within a rectangular matrix (panel~7) to 0.992 for an elliptical void within a double-notch geometry (panel~14). Circular and elliptical voids were reconstructed with high fidelity, consistently achieving IoU values above 0.98 across panels~9--12, 14 and 15, including configurations containing multiple voids. Lobed-star configurations (panels~1--8 and 13) yielded IoU values between 0.911 and 0.966, with errors concentrated primarily around regions of high-curvature concave notches. The lowest reconstruction accuracy was observed for panel~16, in which a U-shaped void embedded within a star geometry achieved an IoU of 0.797 Figure~\ref{fig:2d_obs} extends the observability analysis of
Section~\ref{sec:obs} to all sixteen part-and-void configurations,
evaluating $O(\mathbf{x})$ in the presence of each case's true void.
Consistent with the single-void case in Figure~\ref{fig:obsvoid},
observability is elevated immediately at the void boundary and decays
toward the interior and the clamped edge.

\begin{figure}
\centering
\includegraphics[width=0.79\textwidth]{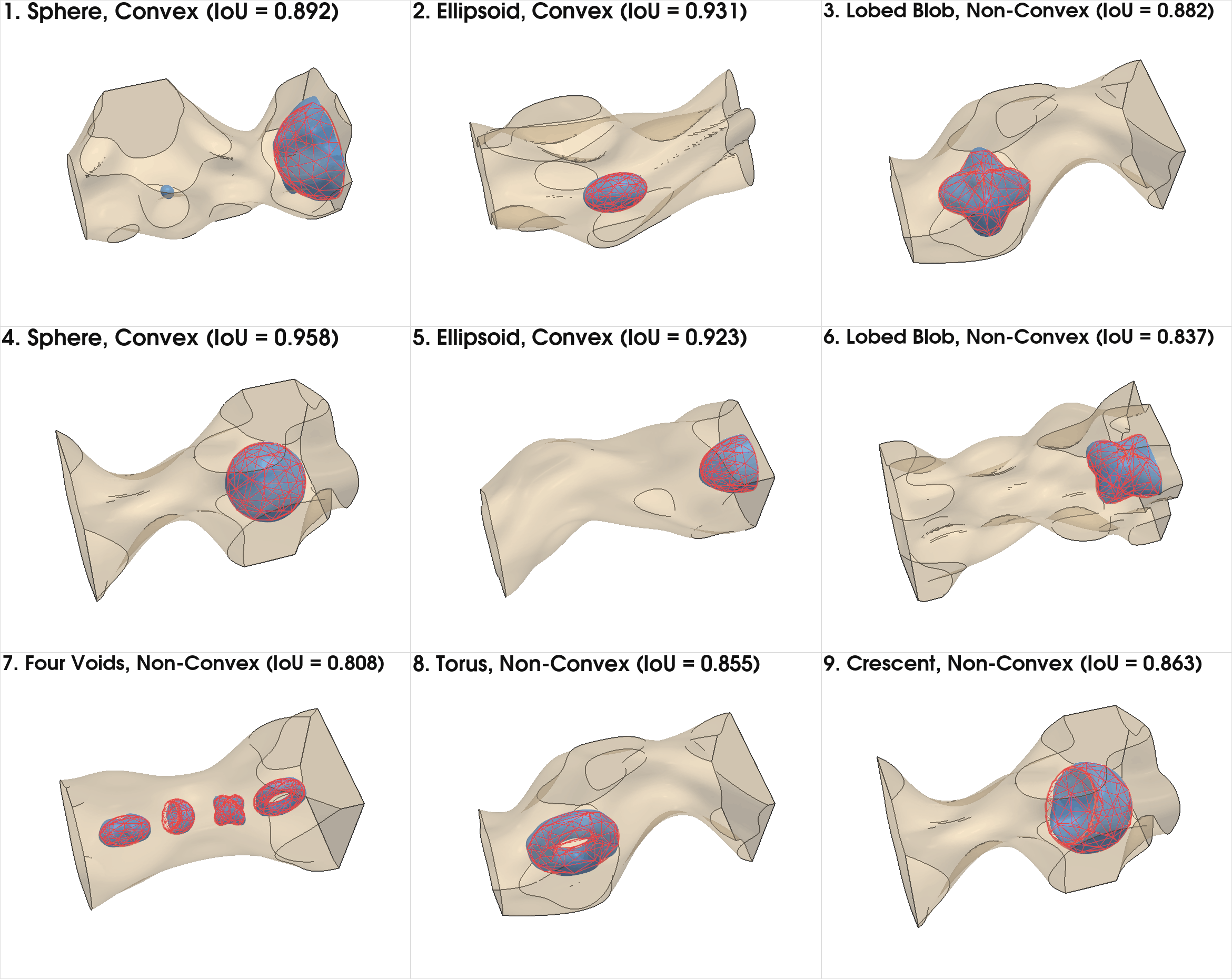}
\includegraphics[width=0.55\textwidth]{figs/fig_legend_void.png}
\caption{Three-dimensional identification of nine voids embedded in arbitrary material geometries, comprising four convex and five non-convex cases. Blue denotes the identified void, the red wireframe the ground-truth void, and the shaded region the surrounding material.}
\label{fig:gallery3d}
\end{figure}

The evaluation was extended to three dimensions using nine voids embedded in
materials with arbitrary geometries (Fig.~\ref{fig:gallery3d}). Four cases
involved convex voids, comprising two spheres (panels~1 and~4;
$\mathrm{IoU}=0.892$ and $0.958$, respectively) and two ellipsoids
(panels~2 and~5; $\mathrm{IoU}=0.931$ and $0.923$, respectively). The
remaining five cases comprised non-convex or multi-void configurations,
including two lobed geometries (panels~3 and~6; $\mathrm{IoU}=0.882$ and
$0.837$), four separate voids (panel~7; $\mathrm{IoU}=0.808$), a torus
(panel~8; $\mathrm{IoU}=0.855$), and a crescent-shaped void (panel~9;
$\mathrm{IoU}=0.863$). Overall, the sphere in panel~4 yielded the highest
reconstruction accuracy ($\mathrm{IoU}=0.958$), whereas the four-void
configuration in panel~7 gave the lowest ($\mathrm{IoU}=0.808$). Importantly,
the toroidal geometry was also reconstructed with better accuracy,
achieving an $\mathrm{IoU}$ of $0.855$.

\begin{figure}[H]
\centering
\includegraphics[width=0.49\textwidth]{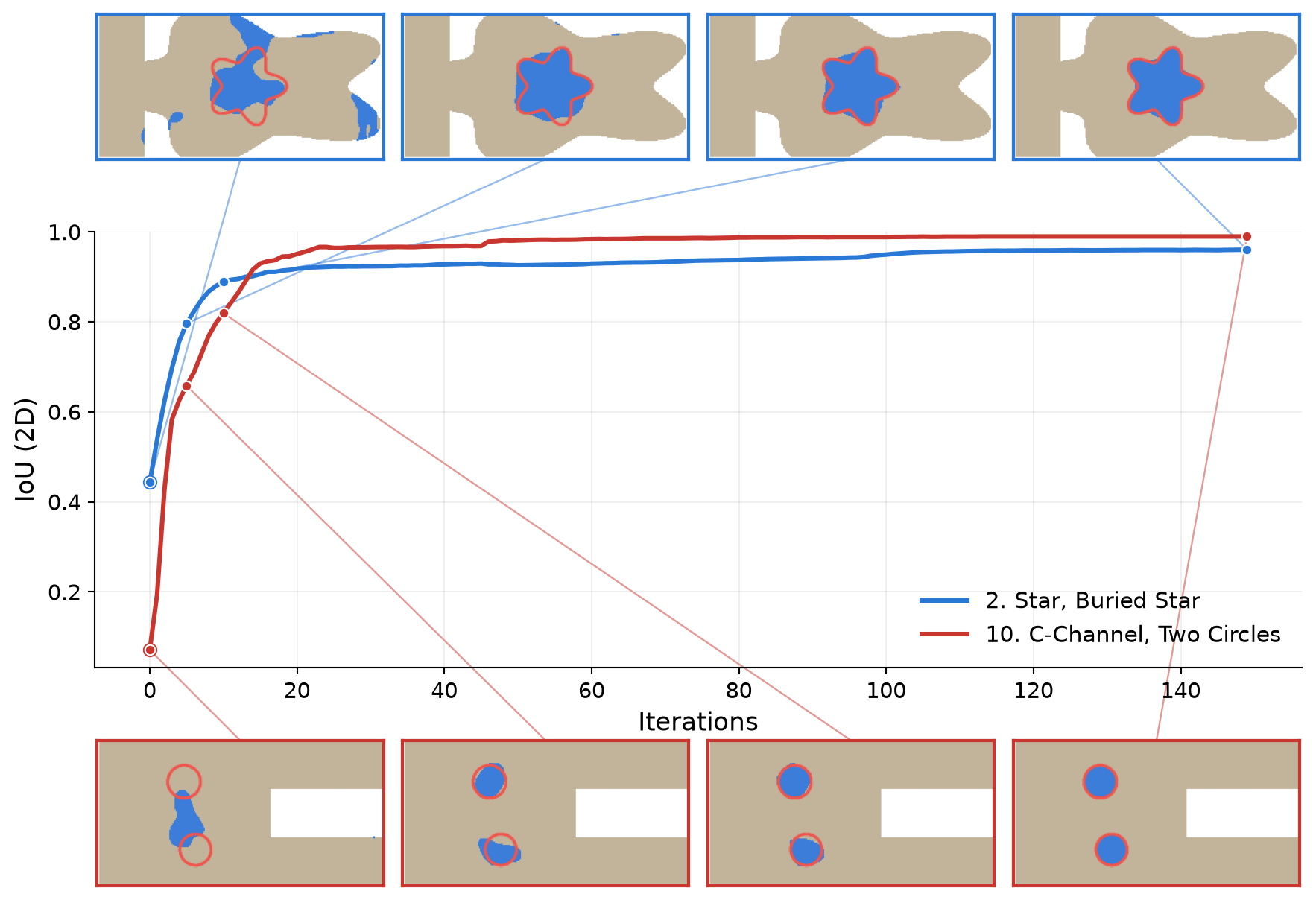}
\includegraphics[width=0.49\textwidth]{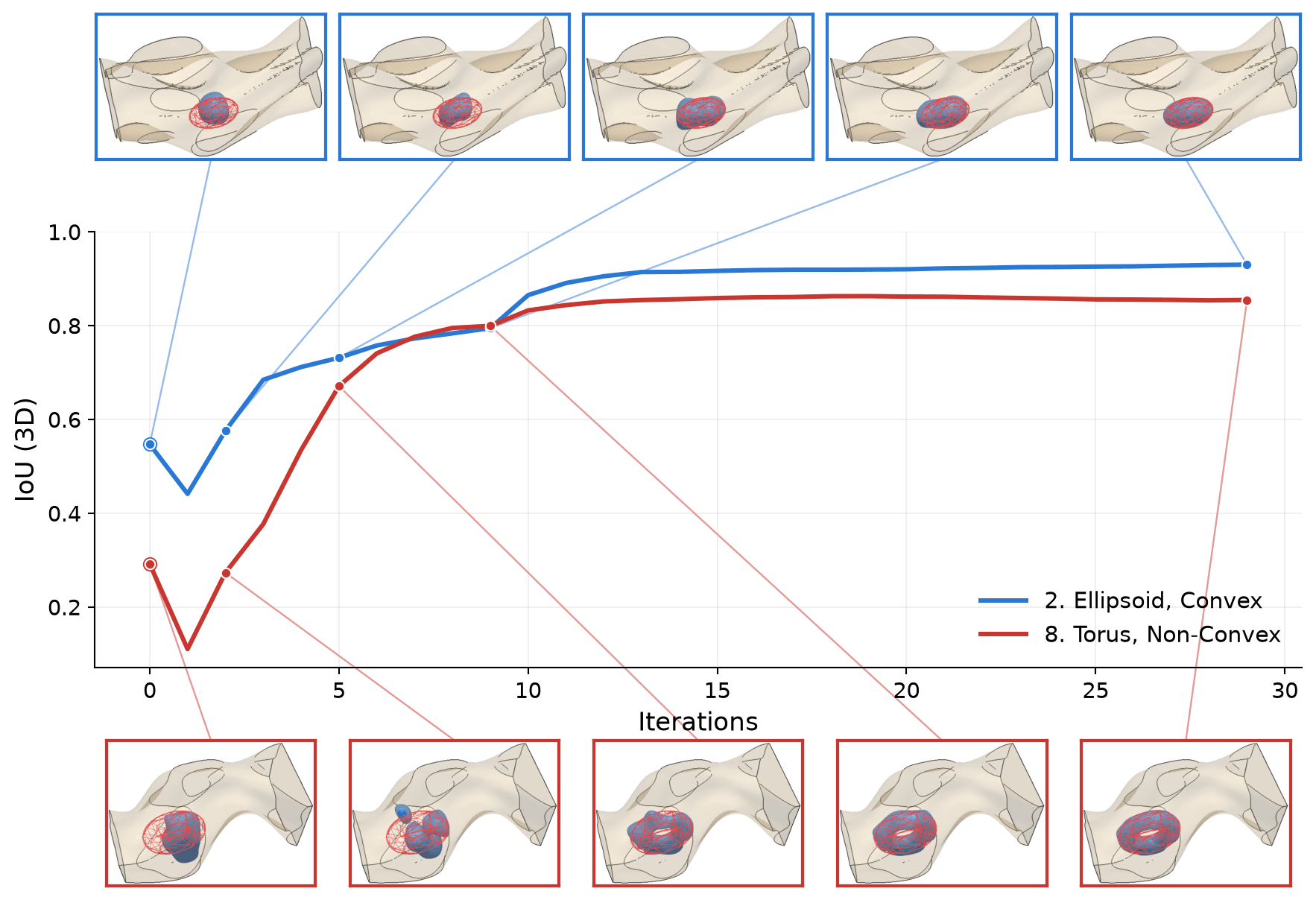}
\caption{Convergence of IoU with the number of LM iterations in two (left) and three (right) dimensions. Iteration 0 denotes the encoder’s initial prediction. Insets show the corresponding reconstructions at the indicated iterations.}
\label{fig:iter}
\end{figure}

Convergence behavior is examined for two representative cases in both two and three dimensions (Fig.~\ref{fig:iter}): the star-shaped void (panel~2 of Fig.~\ref{fig:iter}) and the two circular voids within a C-channel (panel~10) in two dimensions, and the ellipsoidal void (panel~2 of Fig.~\ref{fig:iter}) and torus (panel~8) in three dimensions. At iteration zero, before application of the forward solver, the encoder provides initial IoU values of 0.44 and 0.07 for the two-dimensional cases and 0.55 and 0.29 for the three-dimensional cases, respectively. In the C-channel (two-dimensional) case, the initial estimate incorrectly represents the two distinct voids as a single connected blob.
In two dimensions, both cases exceed $\mathrm{IoU}=0.8$ within ten iterations and reach approximately $0.92$ and $0.97$ by iteration 20. Extending the optimization by a further 130 iterations yields only modest gains, to $\mathrm{IoU}=0.960$ and $0.990$, respectively. In three dimensions, both cases converge within approximately 30 iterations, with the solutions essentially reaching their final values by iteration 15: $\mathrm{IoU}=0.931$ for the ellipsoid and $0.855$ for the torus. Across both dimensions, the overall void geometry is established during the early iterations, whereas subsequent iterations primarily refine the boundary.

The two-dimensional insets show the initial single blob separating into
two circular voids and the star contracting towards its true centre. For
the torus, the central hole opens between iterations~5 and~9, after the
outer envelope is already established. Once these major geometric
features emerge, subsequent iterations primarily refine their boundaries
rather than alter the overall void configuration. The three-dimensional
cases follow the same trend, with the ellipsoid and torus establishing
their essential geometries within the first few iterations, while
subsequent iterations primarily refine the boundaries.

\subsection{Sensor-Count Reduction}
\label{sec:sensors}

The results above use the full set of station measurements ($N_s=256$ in two dimensions and $N_s=200$ in three dimensions). To assess the robustness of the approach to reduced data availability, we systematically decrease the number of stations. Stations are removed sequentially using farthest-point subsampling, which preserves a broad spatial distribution along the boundary. Because the encoder \eqref{eq:setenc} aggregates station measurements symmetrically, it can accommodate varying numbers of stations without retraining.

\begin{figure}[H]
\centering
\includegraphics[width=0.97\textwidth, trim = 0 0.65cm 0 0]{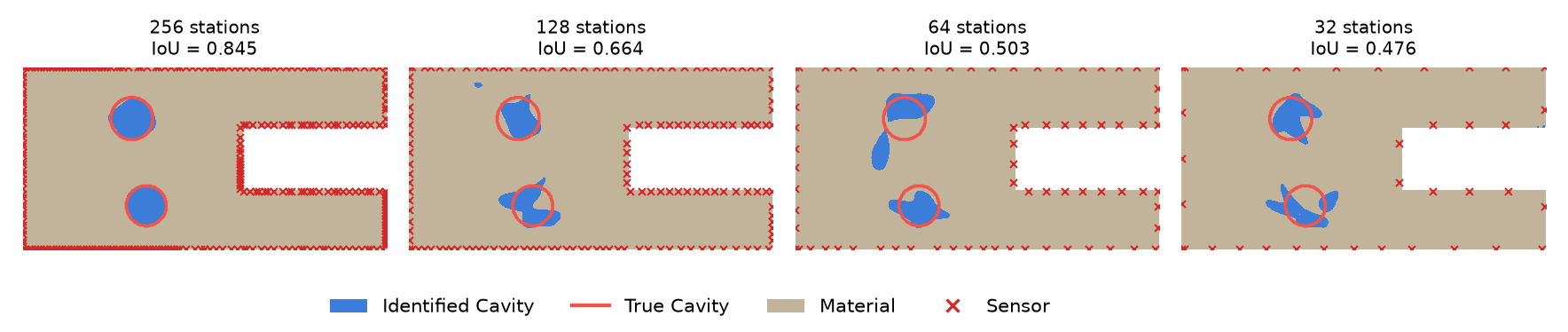}
\includegraphics[width=0.95\textwidth]{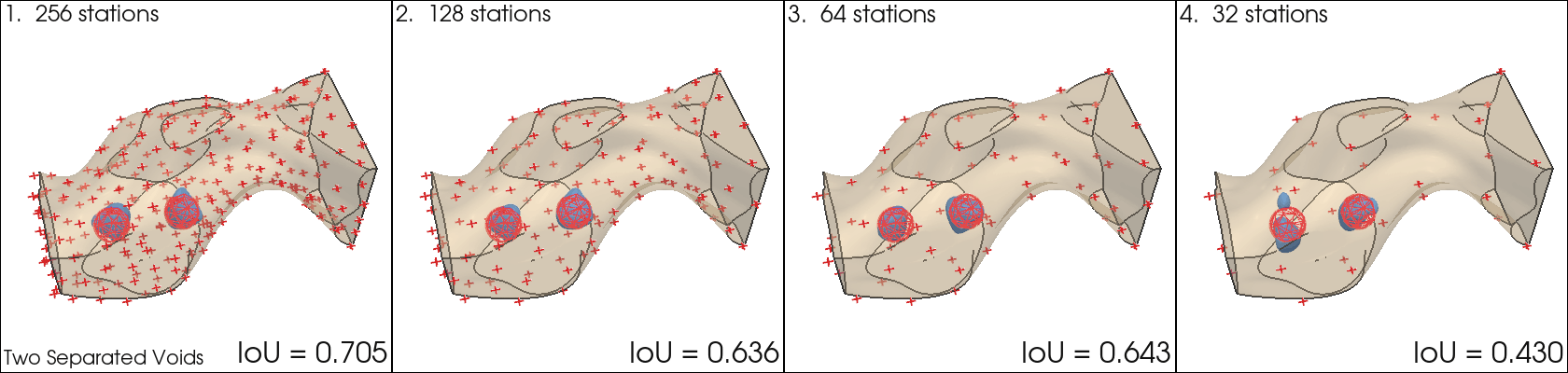}
\includegraphics[width=0.55\textwidth]{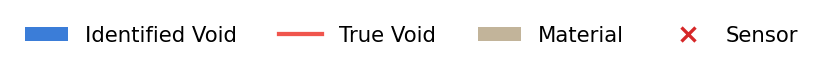}
\caption{Sensor-count reduction. Top, two-dimensional, two-void case. Bottom, three-dimensional, two-void case. Each panel shows IoU against the number of measurement stations. Reconstruction accuracy generally decreases with decreasing station count.}
\label{fig:sens2d3d}
\end{figure}

Reconstruction accuracy, measured by IoU, generally decreases as the number of stations is reduced (Fig.~\ref{fig:sens2d3d}). In two dimensions, the two-void case declines from $\mathrm{IoU}=0.845$ with 256 stations to $0.664$, $0.503$ and $0.476$ with 128, 64 and 32 stations, respectively. In three dimensions, the IoU for two separated voids decreases from $0.705$ with 256 stations to $0.636$, $0.643$ and $0.430$ with 128, 64 and 32 stations, respectively. Although the degradation is not strictly monotonic, the overall trend is consistent across both dimensions. These results demonstrate that station density places a direct constraint on reconstruction accuracy, indicating that the measurement array should be sized according to the accuracy required by the experiment rather than treated as redundant.

\subsection{Uncertainty Quantification with Noisy Measurements}
\label{sec:uq}
To evaluate reconstruction reliability beyond deterministic single-case outputs, we propagate boundary displacement measurement noise through the inversion pipeline to quantify uncertainty via spatially resolved void probabilities. Measurement noise is modeled as additive Gaussian perturbations scaled to the root-mean-square amplitude of the noiseless response:
\begin{equation}
\label{eq:noise}
\mathbf{u}^{0}_{\rm obs} = \mathbf{u}^{0} + \boldsymbol{\epsilon},
\qquad
\boldsymbol{\epsilon} \sim \mathcal{N}\left(0, \sigma^2\mathbf{I}\right),
\qquad
\sigma = \epsilon \sqrt{\overline{\lVert\mathbf{u}^{0}\rVert^2}},
\qquad
\epsilon \in \{1\%, 5\%, 10\% \}.
\end{equation}
Consistency across realizations is maintained by using identical noisy measurements for both encoder input and latent optimization targets. Initialising reconstructions from noiseless priors during noisy optimization would otherwise artificially bias convergence and overestimate robustness.

\begin{figure}
\centering
\includegraphics[width=0.9\textwidth]{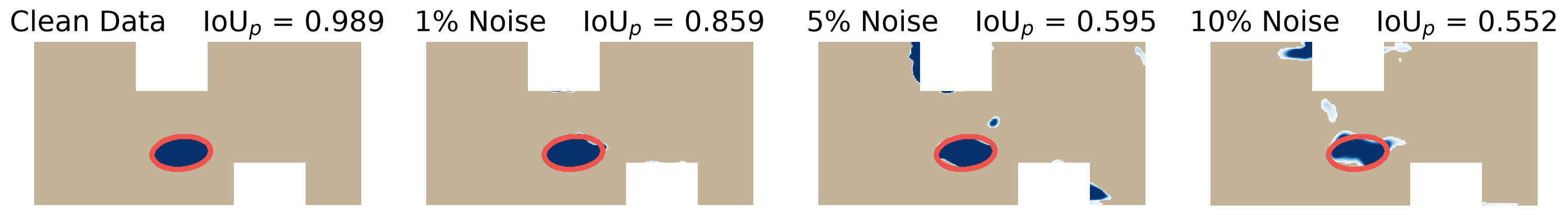}
\includegraphics[width=0.9\textwidth]{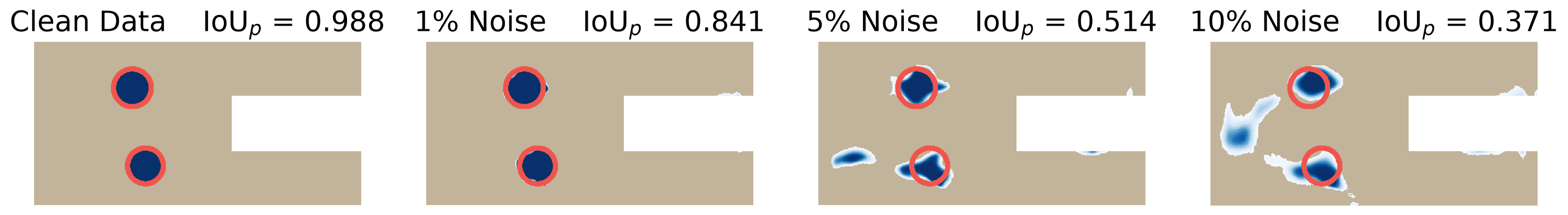}
\includegraphics[width=0.9\textwidth]{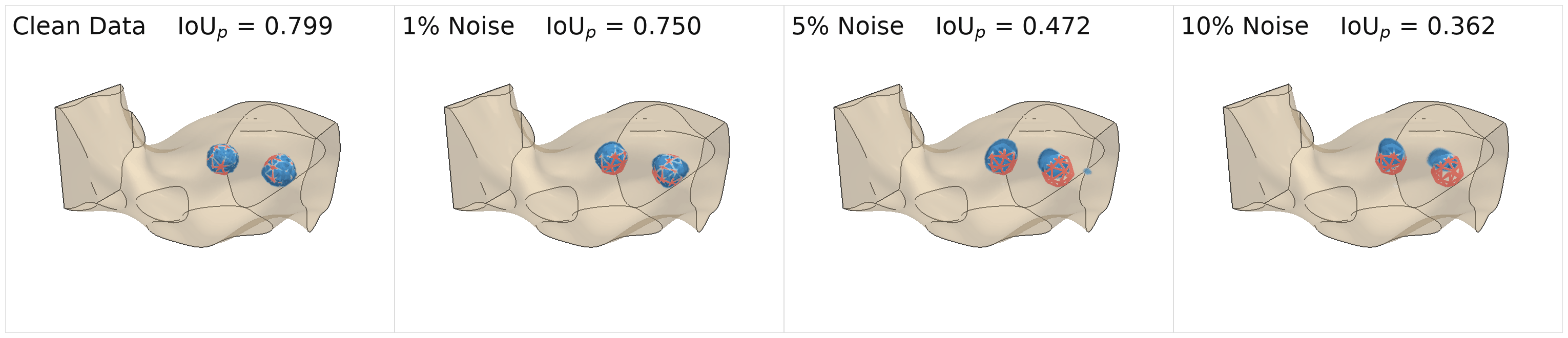}
\includegraphics[width=0.30\textwidth]{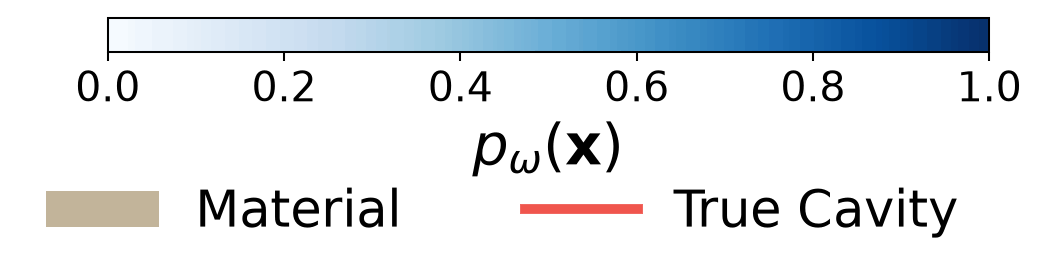}
\caption{Void probability $p_{\omega}(\mathbf{x})$ for three representative test cases. From top to bottom: the two-dimensional single-ellipse case in the double-notch part, the two-dimensional two-void C-channel case, and the three-dimensional two-void case, respectively. At each point, $p_{\omega}(\mathbf{x})\in[0,1]$ denotes the probability that the point belongs to the identified void. White indicates $p_{\omega}=0$, dark blue $p_{\omega}=1$, and intermediate shades denote increasing uncertainty in the reconstructed void boundary. Red line denote the ground-truth voids.}
\label{fig:uq2d3d}
\end{figure}

The two-void C-channel is reconstructed with $\mathrm{IoU}_p=0.988$ on clean data, decreasing to $0.841$, $0.514$ and $0.371$ under $1\%$, $5\%$ and $10\%$ noise, respectively (Figure~\ref{fig:uq2d3d}). Across both two-dimensional cases, the highest-probability regions remain localized at the true voids as noise increases. The principal effect of increasing noise is instead a progressive broadening of diffuse, low-probability regions into the surrounding material, most pronounced at $10\%$ noise. In three dimensions, the two-void reconstruction achieves $\mathrm{IoU}_p=0.799$ on clean data and decreases to $0.750$, $0.472$ and $0.362$ at $1\%$, $5\%$ and $10\%$ noise, respectively, indicating a more pronounced sensitivity to measurement noise.

\subsection{Void Identification in Hyperelastic and Plastic Materials}
\label{sec:nonlinear}
\begin{figure}
\centering
\includegraphics[width=0.85\textwidth]{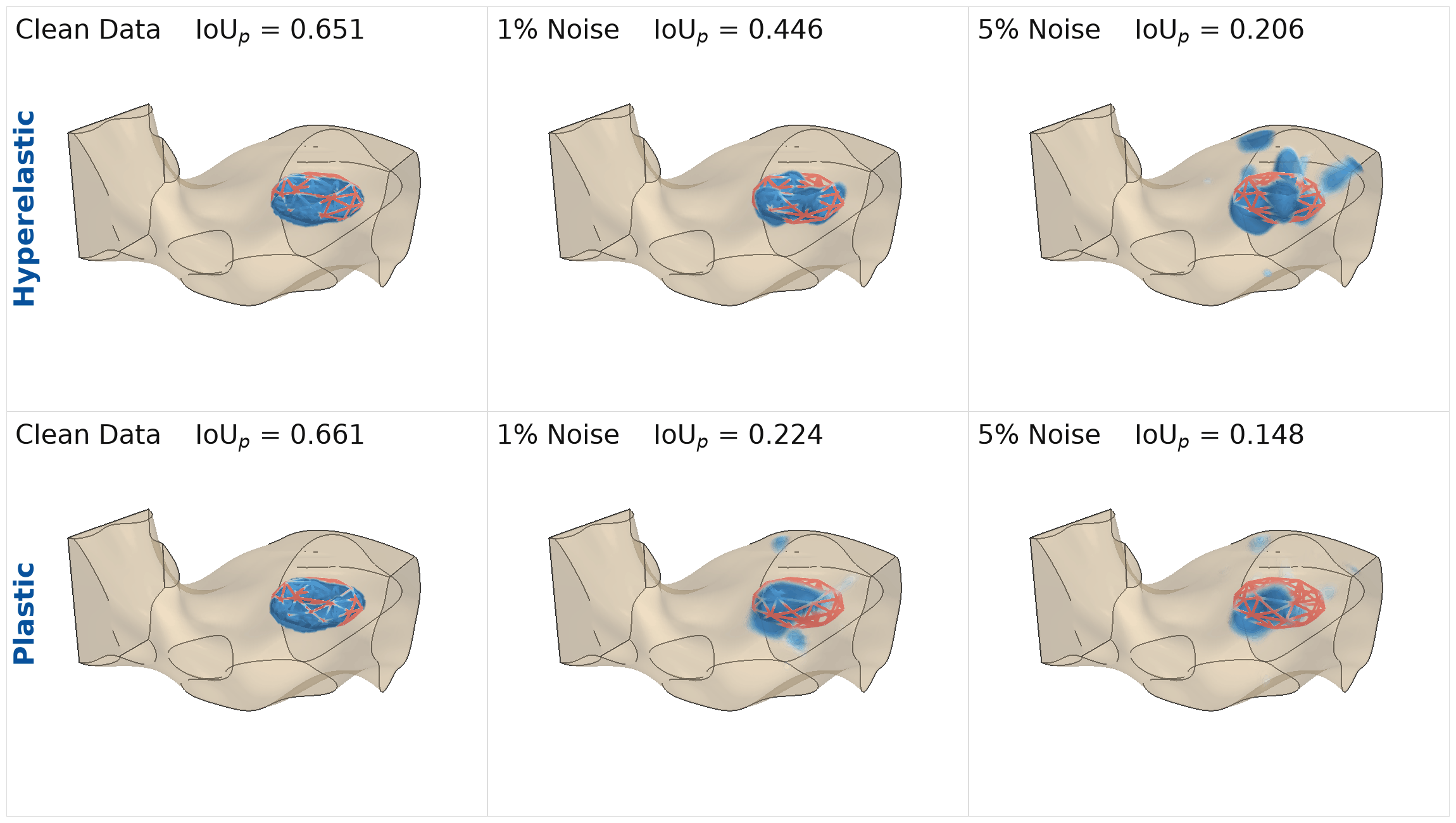}
\includegraphics[width=0.30\textwidth]{figs/fig_uq_key.png}
\caption{Torus void probability $p_{\omega}(\mathbf{x})$ under the hyperelastic (top) and plastic (bottom) constitutive models as a function of measurement noise. For the hyperelastic model, $\mathrm{IoU}_p$ is $0.651$ on clean data and decreases to $0.446$ and $0.206$ at $1\%$ and $5\%$ noise, respectively. For the plastic model, $\mathrm{IoU}_p$ decreases from $0.661$ on clean data to $0.224$ and $0.148$ at $1\%$ and $5\%$ noise, respectively.}
\label{fig:torusnl}
\end{figure}
Beyond linear elasticity, we extend geometry identification to finite deformation and plasticity following the constitutive framework of Zhang et al.~\cite{zhang2022pinn}. For hyperelastic regimes, we employ an incompressible Neo-Hookean solid, in which the first Piola--Kirchhoff stress incorporates an independent pressure field,
\begin{equation}
\label{eq:neo}
\mathbf{P}=\mu\mathbf{F}-p\mathbf{F}^{-\top},
\qquad
\det\mathbf{F}=1,
\end{equation}
with deformation gradient $\mathbf{F}=\mathbf{I}+\nabla_{\mathbf{X}}\mathbf{u}$. To bypass the limitations of displacement-only hexahedral discretizations, this constraint is enforced via a volumetric--deviatoric energy split,
\begin{equation}
\label{eq:neopen}
W(\mathbf{F})
=
\frac{\mu}{2}\left(J^{-2/3}I_1-3\right)
+
\frac{\kappa}{2}(J-1)^2,
\qquad
J=\det\mathbf{F},
\quad
I_1=\mathbf{F}:\mathbf{F},
\end{equation}
recovering $p=-\kappa(J-1)$ in the limit $\kappa/\mu\to\infty$. We set $\kappa/\mu=50$, corresponding to an effective Poisson's ratio of $0.49$, to avoid the artificial numerical stiffening of trilinear hexahedra near incompressibility.

Plasticity is modeled using compressible deformation plasticity with power-law hardening. Because the experimental setup involves strictly monotone proportional loading without load reversal, we adopt the equivalent Ramberg--Osgood formulation implemented in Abaqus. This nonlinear-elastic approximation avoids path-dependent sensitivities through incremental load steps, keeping the inverse problem computationally tractable. The deviatoric response is governed by the von Mises stress $q$, equivalent strain $e_{\rm eq}$, deviatoric strain $\mathbf{e}$, yield stress $\sigma_0$, hardening exponent $n=5$, and offset $\tau=3/7$:
\begin{equation}
\label{eq:ro}
e_{\rm eq}
=
\frac{q}{3G}
+
\tau\frac{q}{E}
\left(\frac{q}{\sigma_0}\right)^{n-1},
\qquad
\mathbf{s}
=
\frac{2q}{3e_{\rm eq}}\mathbf{e},
\qquad
p=\kappa\,\mathrm{tr}\,\boldsymbol{\varepsilon},
\end{equation}
where the second term vanishes below yield, reducing the formulation exactly to linear elasticity.

For the nonlinear material studies, we consider the torus void and evaluate reconstruction performance under hyperelastic and plastic constitutive models as a function of measurement noise (Figure~\ref{fig:torusnl}). On clean data, the two models yield comparable accuracy, with $\mathrm{IoU}_p=0.651$ for the hyperelastic model and $0.661$ for the plastic model. With increasing noise, however, their performance diverges. For the hyperelastic model, $\mathrm{IoU}_p$ decreases to $0.446$ at $1\%$ noise and $0.206$ at $5\%$, whereas the plastic model exhibits a sharper decline, reaching $0.224$ and $0.148$, respectively. In both cases, the highest-probability region remains localized near the true void at low noise, while increasing noise produces diffuse uncertainty in the surrounding material.

\subsection{Experimental Validation of Void Reconstruction}\label{sec:EXP}
\begin{figure}
    \centering
    \includegraphics[scale=1,clip=true]{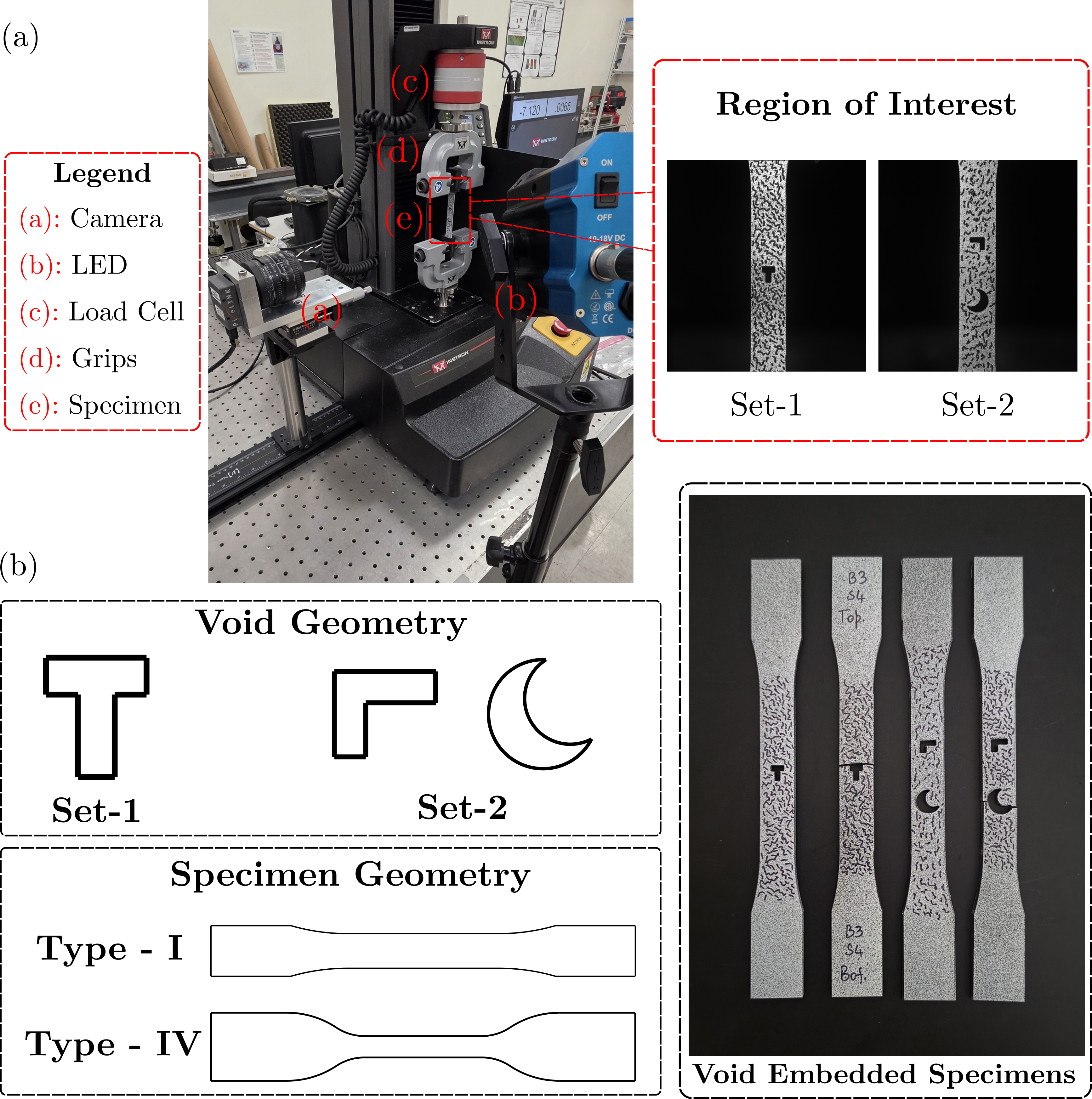}
    \caption{Schematic illustration of the void and specimen geometries and uniaxial tensile experimental setup used in this study. (a) Uniaxial tensile experiments are performed on speckle-patterned, void embedded specimens and the boundary displacement profiles are recorded using 2D DIC. (b) Three void shapes, T-shaped, L-shaped and crescent-shaped are grouped into two sets to illustrate the GenVoid's ability to quantify the presence, shape and location of arbitrarily shaped voids. The voids are embedded within the gage region of ASTM standard specimens. These specimens are shown before and after fracture for the elastoplastic material. }
    \label{fig:11}
\end{figure}

We experimentally validate GenVoid using physical test specimens containing voids with arbitrarily prescribed geometries. Two distinct void configurations, shown in Figure~\ref{fig:11}, were embedded within ASTM-standard tensile specimens and subjected to uniaxial tensile loading. Full-field displacements were measured using DIC and subsequently provided as input to the trained GenVoid surrogate for void reconstruction. The reconstructed geometries were then compared with the corresponding experimentally prescribed voids to assess the fidelity and robustness of the surrogate under realistic experimental conditions. Detailed descriptions of the void-geometry design, specimen fabrication, experimental setup and testing protocol are provided in Appendix~\ref{exp}.

\begin{figure}
\centering
\includegraphics[scale=0.18,clip=true]{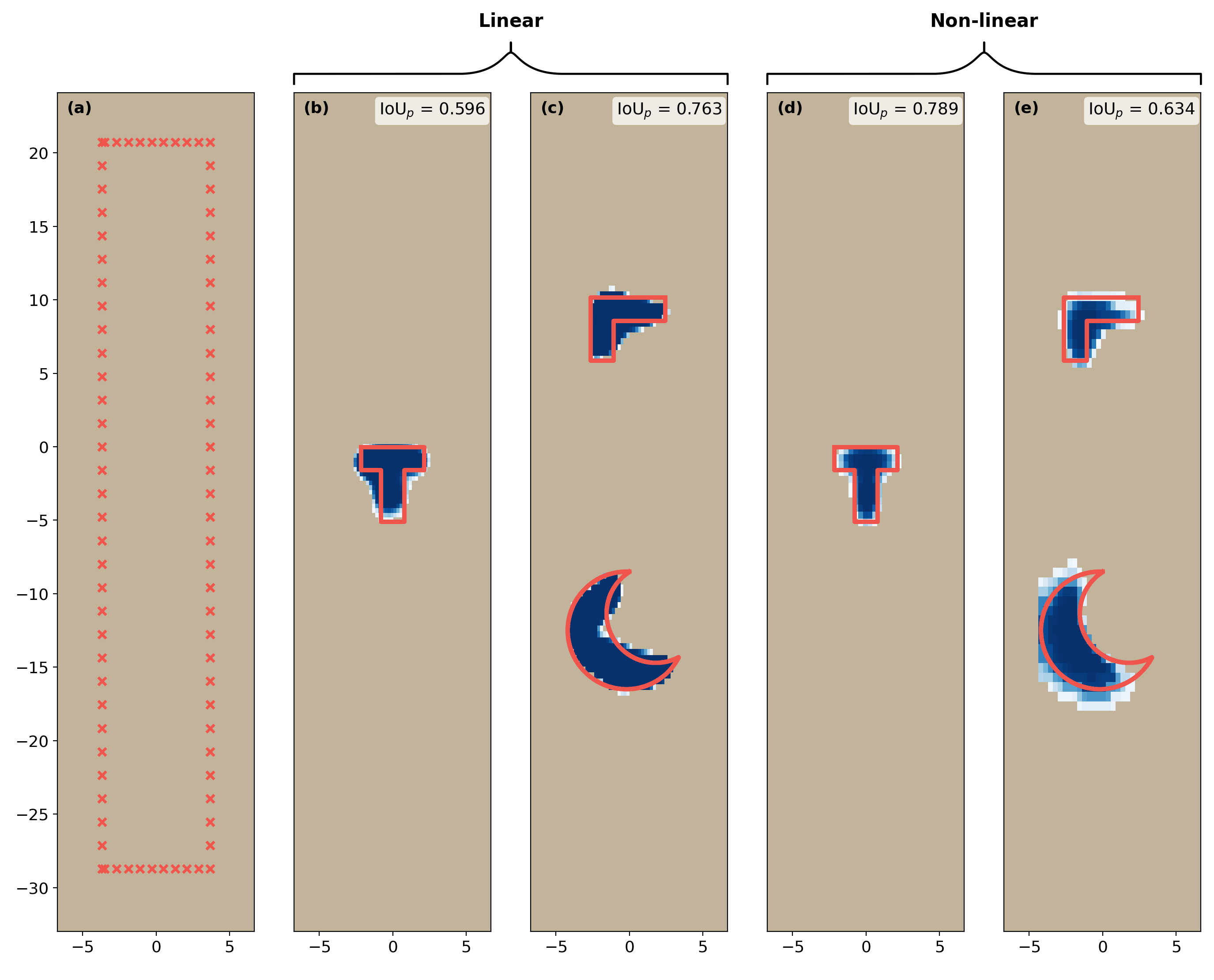}

\includegraphics[scale=0.18,clip=true]{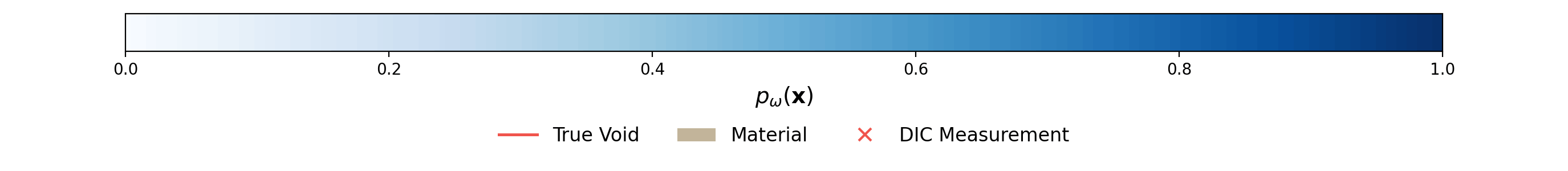}

\caption{Void identification from experimental DIC measurements in linear-elastic (PMMA) and elastoplastic specimens. (a) Specimen boundary and DIC-derived sensor locations used for all reconstructions. (b) Void probability $p_{\omega}(\mathbf{x})$ for the Set-1 T-shaped void under linear-elastic behaviour, with $\mathrm{IoU}p = 0.596$. (c) Void probability $p_{\omega}(\mathbf{x})$ for the Set-2 specimen, containing L-shaped and crescent-shaped voids, under linear-elastic behaviour, with $\mathrm{IoU}p = 0.763$. (d) Void probability $p_{\omega}(\mathbf{x})$ for the Set-1 T-shaped void under elastoplastic behaviour, with $\mathrm{IoU}p = 0.789$. (e) Void probability $p_{\omega}(\mathbf{x})$ for the Set-2 specimen under elastoplastic behaviour, with $\mathrm{IoU}_p = 0.634$. The true void boundary is shown by the continuous red line.}
\label{fig:EXP}
\end{figure}

We first evaluate GenVoid on linear-elastic specimens using experimentally measured boundary displacement fields and an acrylic (PMMA) matrix, and then assess its performance on elastoplastic specimens, for which the matrix response is described by $J_2$ deformation theory \cite{hutchinson1981finite} with isotropic hardening. The DIC measurements provide a direct experimental basis for evaluating the fidelity of the inferred void geometries. Figure~12 compares GenVoid reconstructions for the Set-1 and Set-2 specimens with their corresponding ground-truth geometries under both linear-elastic and elastoplastic matrix behaviour. For each reconstruction, the void probability field is computed from $N=1000$ posterior samples. Figure~12a shows the specimen boundary and sensor locations extracted from the DIC displacement field; this sensor layout is common to all four reconstructions. For the Set-1 specimen, the T-shaped void is recovered with $\mathrm{IoU}_p = 0.596$ under linear-elastic behaviour (Fig.~12b) and $\mathrm{IoU}_p = 0.789$ under elastoplastic behaviour (Fig.~12d). In both cases, the predicted probability field is concentrated within the true void boundary. For the Set-2 specimen, which contains L-shaped and crescent-shaped voids, both voids are correctly identified as separate, disconnected regions under both matrix behaviours, with a combined $\mathrm{IoU}_p = 0.763$ under linear-elastic behaviour (Fig.~12c) and $\mathrm{IoU}_p = 0.634$ under elastoplastic behaviour (Fig.~12e). The elastoplastic reconstruction exhibits a more diffuse probability field around the crescent-shaped void than the corresponding linear-elastic reconstruction.

\section{Conclusions}\label{sec:conclusions}
Characterizing internal defects from accessible measurements remains a fundamental challenge in non-destructive evaluation and structural health monitoring because, unlike surface damage, internal voids are not directly observable and their signatures at the boundary depend on defect location, structural geometry, material response and applied loading. This renders the inverse problem intrinsically ill-posed, as different internal configurations can produce similar surface measurements and defects in poorly observable regions may leave only weak signatures in the measured response. Beyond recovering defect geometry, accurate reconstruction establishes a direct link between the location and morphology of internal voids and their influence on material behaviour, providing a basis for quantifying local stress concentrations, heat-transfer pathways and electrical transport, and ultimately for assessing their effects on mechanical failure, thermal conductivity and electronic performance. Here, we introduce \textit{GenVoid}, a physics-informed generative model that reconstructs hidden void geometries from surface displacement measurements by representing candidate geometries in a compact latent space and evaluating them through a governing finite-element model during inference. The main findings of this study are:

\begin{itemize}
\item The  GenVoid enables reconstruction of diverse two- and three-dimensional defect morphologies, including multiple voids, while accommodating different material responses ranging from linear elasticity to hyperelasticity and plasticity.

\item The probabilistic latent formulation provides a means of propagating measurement uncertainty into uncertainty in the reconstructed geometry, allowing the inference of a distribution of plausible defect configurations rather than a single deterministic solution.

\item Reconstructions from both high-fidelity synthetic data and experimentally measured displacement fields demonstrate that hidden defects can be identified from boundary measurements, including in the presence of measurement noise and nonlinear material behaviour.

\item The observability analysis establishes a quantitative connection between defect location and the information available in boundary measurements. By evaluating the sensitivity of boundary responses to localized interior perturbations under multiple loading conditions, it identifies regions that are more or less constrained by the measurement system.
This observability information can be used to guide experimental design by assessing the effects of sensor number, placement and loading conditions on the information available for defect reconstruction.

\end{itemize}

More broadly, these results demonstrate the potential of inverse void identification to connect internal defect geometry with macroscopic functional and mechanical behaviour, while shifting hidden-defect identification from a purely reconstructive task towards an integrated problem of inference, uncertainty quantification and measurement design. By establishing where and with what confidence boundary sensing can resolve internal voids, this framework provides a basis for designing more efficient inspection strategies and for assessing the reliability of inferred defect geometries and their associated material responses.

The present framework assumes that the constitutive behaviour and material properties of the host material are known, which isolates the geometric inverse problem but may limit its applicability to heterogeneous and composite systems, where spatial variations in material properties can also shape the measured boundary response. A natural extension is to jointly infer defect geometry and material properties within the same physics-informed framework, including phase-dependent properties, anisotropy and spatially varying constitutive parameters. Such extensions could also accommodate broader defect morphologies and more complex internal structures beyond the geometric distributions represented during training. More generally, integrating experimentally calibrated constitutive behaviour with richer sensing modalities, multiple loading conditions and adaptive loading or sensor placement could enable joint geometry–material inference and further extend \textit{GenVoid} towards non-invasive characterization of complex materials and structures.

\appendix
\section{Nomenclature}
\label{sec:nomenclature}
\small
\renewcommand{\arraystretch}{0.9}
\setlength{\tabcolsep}{4.0pt}

\noindent\begin{minipage}[t]{0.49\textwidth}
\textbf{Materials and Forward Problem}\\[2pt]
\begin{tabular}{@{}ll@{}}
\toprule
Symbol & Definition \\
\midrule
$\Omega_0$ & Reference domain \\
$\Gamma_u,\ \Gamma_t$ & Dirichlet, Neumann boundary \\
$\omega$ & Unknown interior void \\
$\Gamma_w$ & Void boundary \\
$\Omega_\omega$ & Void domain \\
$\Omega$ & Material domain ($\Omega_0 \setminus \Omega$ ) \\
$\mathbf{u}$ & Displacement field \\
$\mathbf{P}(\mathbf{u})$ & First Piola--Kirchhoff stress \\
$\mathbf{b}_0$ & Body force \\
$\hat{\mathbf{n}},\ \hat{\mathbf{n}}_w$ & Outward normal, outer/void bdry.\ \\
$\mathbf{u}_D,\ \mathbf{t}_N$ & Prescribed displacement, traction \\
$\boldsymbol{\sigma}$ & Cauchy stress tensor \\
$\mathbb{C}$ & Elasticity (stiffness) tensor \\
$\boldsymbol{\varepsilon}$ & Linearized strain tensor \\
$\mathbf{F}$ & Deformation gradient \\
$W(\mathbf{F})$ & Strain energy density \\
$\mathbf{s},\ \mathbf{e}$ & Deviatoric stress, strain \\
$q$ & Von Mises equivalent stress \\
$p$ & Hydrostatic pressure \\
$\kappa$ & Bulk modulus \\
$\sigma_0$ & Reference yield stress \\
$n$ & Strain hardening exponent \\
$\tau$ & Ramberg--Osgood yield offset \\
$G$ & Shear modulus (Ramberg--Osgood) \\
$\mu$ & Shear modulus (Neo-Hookean) \\
$J=\det\mathbf{F}$ & Jacobian of deformation \\
$I_1=\mathbf{F}\!:\!\mathbf{F}$ & First deformation invariant \\
$\mathbf{R}(\mathbf{u}_h;\omega)$ & Global nonlinear residual \\
$\mathbf{F}_{\mathrm{int}},\ \mathbf{F}_{\mathrm{ext}}$ & Internal, external force vectors \\
$\mathcal{O}_{\mathrm{ad}}$ & Admissible set of void topologies \\
$\mathbf{u}_h$ & FEM-predicted displacement \\
$\mathbf{u}^0$ & Measured (observed) displacement \\
$N_s,\ N_l$ & \# Sensor stations, load cases \\
\bottomrule
\end{tabular}
\end{minipage}
\hfill
\noindent\begin{minipage}[t]{0.49\textwidth}
\textbf{Level-Set Representation}\\[2pt]
\begin{tabular}{@{}ll@{}}
\toprule
Symbol & Definition \\
\midrule
$\phi(\mathbf{x})$ & Level-set field, void where $\phi<0$ \\
$\mathcal{X}$ & Boundary geometry coordinates \\
$\mathbf{a}$ & RBF coefficient vector \\
$a_j$ & $j$th RBF coefficient \\
$N_R$ & Total number of RBF knots \\
$g_j(\mathbf{x})$ & $j$th RBF basis function \\
$\mathbf{x}_j$ & Center of $j$th RBF knot \\
$\delta$ & RBF kernel support-scale parameter \\
$c$ & RBF kernel regularization constant \\
$N_{kx},N_{ky},N_{kz}$ & Knot-grid resolution per axis \\
$\mathcal{G}$ & Node-to-knot interpolation matrix \\
$E_e(\mathbf{a})$ & Effective element Young's modulus \\
$E_0$ & Young's modulus of solid material \\
$H_b(\phi)$ & Smoothed Heaviside interpolation \\
$\gamma$ & Ersatz stiffness floor \\
$b,\ b_0$ & Interface width, its nominal value \\
$\mathbf{K}(\mathbf{a})$ & Global stiffness matrix \\
$\mathbf{K}_0^{(e)}$ & Reference element stiffness matrix \\
\bottomrule
\end{tabular}

\vspace{19pt}
\textbf{Observability}\\[2pt]
\begin{tabular}{@{}ll@{}}
\toprule
Symbol & Definition \\
\midrule
$O(\mathbf{x})$ & Observability field \\
$E(\mathbf{x})$ & Local Young's modulus \\
$\mathbf{u}_{h,l}$ & Predicted response, load case $l$ \\
$O_{\max}$ & Maximum observability value \\
$\mathrm{Vol}$ & Void volume \\
$\Delta\mathbf{u}$ & Measurement footprint \\
\bottomrule
\end{tabular}
\end{minipage}

\vspace{8pt}
\noindent\begin{minipage}[t]{\textwidth}
\textbf{Neural Network Architecture and Training}\\[2pt]
\begin{tabular}{@{}ll@{}p{0.5cm}ll@{}}
\toprule
Symbol & Definition && Symbol & Definition \\
\midrule
$\mathbf{z}$ & Latent vector && $\phi_1(\mathbf{z})$ & Decoded (prior) level-set field \\
$d$ & Latent dimension && $\boldsymbol{\xi}_s$ & Per-station feature vector \\
$\mathcal{E}_{\boldsymbol{\theta}}$ & Encoder && $\mathbf{X}_s$ & Station spatial coordinates \\
$\mathcal{D}_{\boldsymbol{\psi}}$ & Decoder &&& \\
$\mathcal{H}_{\tilde{\boldsymbol{\theta}}}$ & MLP-head network && $\boldsymbol{\alpha}(\mathbf{z})$ & Correction field, $\mathcal{G}\mathcal{H}_{\tilde{\boldsymbol{\theta}}}(\mathbf{z})$ \\
$\boldsymbol{\theta},\boldsymbol{\psi},\tilde{\boldsymbol{\theta}}$ & Encoder, decoder, head weights && $\boldsymbol{\Theta}$ & Opt.\ variable, $\{\mathbf{z},\tilde{\boldsymbol{\theta}}\}$ \\
$\boldsymbol{\mu},\ \boldsymbol{\sigma}$ & Latent posterior mean, std.\ dev.\ && $N$ & Number of posterior samples \\
$\boldsymbol{\epsilon}$ & Noise && $p_{\omega}(\mathbf{x})$ & Void probability field \\
$q_k(\mathbf{z})$ & Latent posterior (training shape $k$) && $y(x)$ & Ground-truth void indicator \\
$D_{\mathrm{KL}}$ & Kullback--Leibler divergence && $\mathrm{IoU}_p,\ \mathrm{IoU}$ & Soft, hard intersection-over-union \\
$\beta$ & KL regularization weight && $\mathcal{J} (\boldsymbol{\Theta})$ & Loss function \\
$\phi^{(k)}$ & Target level-set field (shape $k$) && $\lambda$ & Tikhonov regularization weight \\
$\mathcal{L}_{\mathrm{dec}},\ \mathcal{L}_{\mathrm{enc}}$ & Decoder, encoder training loss &&& \\
\bottomrule
\end{tabular}
\end{minipage}

\vspace{0.2cm}
\section{Experimental Design}\label{exp}
\subsection{Geometric Design of Voids}
Mathematically, the T-void shape specified in Set-1 can be represented using an implicit function, $\varepsilon_T(\mathbf{p})$ which is given by:

\begin{equation}
    \varepsilon_T(\mathbf{p}) = \min \left( \varepsilon_{\text{bar}}(\mathbf{p}'), \varepsilon_{\text{stem}}(\mathbf{p}') \right)
\end{equation}

where the top bar and vertical stem are defined using max-norm bounding boxes:

\begin{equation*}
    \varepsilon_{\text{bar}}(\mathbf{p}') = \max\left(\vert{}x'\vert{} - \frac{w}{2}, \left\vert{}y' - \frac{\tilde{h} - t_h}{2}\right\vert{} - \frac{t_h}{2}\right)
\end{equation*}

and,

\begin{equation*}
    \varepsilon_{\text{stem}}(\mathbf{p}') = \max\left(\vert{}x'\vert{} - \frac{t_v}{2}, \left\vert{}y' + \frac{t_h}{2}\right\vert{} - \frac{\tilde{h} - t_h}{2}\right)
\end{equation*}

Here $\mathbf{p}' = (x',y')$ denotes the local coordinates, translated and rotated from the global coordinates $\mathbf{p}$. Similarly, the L-shaped void in Set-2 is formed by the simple geometric union of vertical and horizontal rectangles. This void shape can also be represented by an implicit function $\varepsilon_L(\mathbf{p})$:

\begin{equation}
    \varepsilon_L(\mathbf{p}) = \min \left( \varepsilon_{\text{vert}}(\mathbf{p}'), \varepsilon_{\text{horiz}}(\mathbf{p}') \right)
\end{equation}

where the horizontal bar ($\varepsilon_{\text{horiz}}(\mathbf{p}')$) and vertical bar ($\varepsilon_{\text{vert}}(\mathbf{p}')$) are defined as:

\begin{equation*}
    \varepsilon_{\text{horiz}}(\mathbf{p}') = \max\left(\vert{}x'\vert{} - \frac{w}{2}, \left\vert{}y' + \frac{\tilde{h} - t_h}{2}\right\vert{} - \frac{t_h}{2}\right)
\end{equation*}

and,

\begin{equation*}
    \varepsilon_{\text{vert}}(\mathbf{p}') = \max\left(\left\vert{}x' + \frac{w - t_v}{2}\right\vert{} - \frac{t_v}{2}, \left\vert{}y' - \frac{t_h}{2}\right\vert{} - \frac{\tilde{h} - t_h}{2}\right)
\end{equation*}

Finally, the crescent-shaped void in Set-2 is generated via the geometric difference between two disks. Specifically, a secondary inner disk (defined by center $\mathbf{c}_2$ and radius $R_2$) is subtracted from a primary outer disk, defined by center $\mathbf{c}_1$ and radius $R_1$. The resulting crescent is defined by the implicit function $\varepsilon_{\text{cres}}(\mathbf{p})$:

\begin{equation}
    \varepsilon_{\text{cres}}(\mathbf{p}) = \max \Big( \Vert{}\mathbf{p} - \mathbf{c}_1\Vert{} - R_1, \; -(\Vert{}\mathbf{p} - \mathbf{c}_2\Vert{} - R_2) \Big)
\end{equation}

where $\Vert{}\cdot\Vert{}$ denotes the Euclidean norm. 

Following their geometric definition, both sets of voids were embedded into ASTM D638 specimens. To thoroughly evaluate the surrogate model's performance across a diverse range of mechanical behaviors, three candidate constitutive materials were selected to represent the linear-elastic, hyperelastic, and elastoplastic behavior in the surrounding matrix. This is detailed in the next section.

\subsection{Test Specimen Fabrication}
Test specimens containing Set-1 and Set-2 voids were fabricated by using conventional and additive manufacturing techniques. To represent a linear-elastic constitutive response in the 2D rectangular gage, a brittle acrylic polymer was selected as the base material. Clear, cast acrylic sheets (12 in.$ \times$ 12 in.$ \times$ 1/8 in.) were obtained from McMaster-Carr (Robbinsville, NJ) and cut into ASTM D638 Type I specimen geometries using a Full Spectrum P-series industrial laser cutter. The Set-1 and Set-2 void geometries were embedded into the Type I specimen matrix using Python, and the resulting profiles were uploaded as vector graphics paths into RetinaEngrave 5 laser preparation software. Specimens were cut by operating the laser at 30\% of its maximum speed and 70\% of its maximum power; two passes were required to fully separate the specimens from the polymethyl methacrylate (PMMA) sheet. The final specimens were allowed to cool at room temperature so as to remove residual stresses and coated with a black and white speckle pattern using flat spray paint prior to tensile testing.

Test specimens exhibiting hyperelastic and elastoplastic matrix behaviors were fabricated via resin-based additive manufacturing on a Formlabs Form 4 3D printer. The Form 4 employs a masked stereolithography (MSLA) process to cure the resin, successfully resolving micron-level features, an important requirement to minimize the influence of process-induced defects in the void-embedded specimens. Both Type I and Type IV specimens were printed using Formlabs' Durable and Flexible 80A resins to replicate the targeted elastoplastic and hyperelastic constitutive behaviors, respectively. Specimens were fabricated flat on the build platform with a layer thickness of 100 $\mu$m, washed with isopropyl alcohol (IPA), and UV-cured post-printing. Consistent with the PMMA specimens, all 3D-printed specimens were coated with a flat black-and-white speckle pattern prior to tensile testing.
The nominal dimensions of the test specimens are detailed in Table~\ref{tab:tab1}.

\begin{table}[htbp]
    \centering
    \caption{Nominal dimensions of Type I and Type IV void-embedded test specimens.}
    \label{tab:tab1}
    \begin{tabular}{@{}llcc@{}}
        \toprule
        \textbf{Parameter} & \textbf{Description} & \textbf{Type I (mm)} & \textbf{Type IV (mm)} \\
        \midrule
        $W$ & Gage width & 13.0 & 6.0 \\
        $L$ & Gage length (height) & 57.0 & 33.0 \\
        $W_{tab}$ & Tab width (overall width) & 19.0 & 19.05 \\
        $L_{tab}$ & Tab height & 30.0 & 21.25 \\
        $R$ & Fillet radius & 76.0 & 14.0 \\
        $T$ & Depth (thickness) & 3.175 & 3.2 \\
        \bottomrule
    \end{tabular}
\end{table}

\subsection{Quasi-Static Tensile Testing}
Uniaxial tensile testing on the void embedded test specimens was conducted using an Instron 5944 universal testing machine equipped with a 2 kN load cell. The testing protocols were specifically tailored to the specimen geometries and the constitutive behaviors of the matrix materials. 

For the Type I specimens, a monotonic loading protocol was employed. These specimens were pulled in tension at a constant crosshead displacement rate of 5 mm/min. Tests were conducted until the specimen failed. Conversely, the Type IV specimens, used to evaluate the hyperelastic matrix behavior, underwent a cyclic loading regimen to capture a stabilized material response. In particular, hyperelastic specimens were subjected to at least 5 continuous loading-unloading cycles up to a peak applied strain of 10\%, also utilizing a crosshead speed of 5 mm/min to eliminate viscoelastic effects. In an effort to minimize the influence of the Mullins effect and extract a representative hyperelastic response, only the force and displacement data recorded during the final loading cycle were utilized for subsequent analysis.

The proposed 2D neural surrogate is trained exclusively on boundary displacement data. To achieve this, the camera was focused on the void-embedded gage region of the specimen during loading and full-field kinematics in the gage region were measured using two-dimensional digital image correlation (2D-DIC). The optical setup consisted of a 5-Megapixel Pixelink digital camera capturing images at a resolution of 2144 $\times$ 2280 pixels. High-intensity NILA LED lighting was utilized to uniformly illuminate the specimens, providing high contrast for the previously applied black-and-white speckle patterns while minimizing exposure times. The captured image sequences were then imported into VIC-2D 7 software (Correlated Solutions, Inc.), which was used to compute and map the full-field in-plane displacement fields across the testing domain throughout the loading history. The experimental setup is illustrated in Figure~\ref{fig:11}(b). 

\bibliographystyle{elsarticle-num} 
\bibliography{reference}
\end{document}